\documentclass[preprint,sort,compress,12pt]{elsarticle}

\usepackage{amssymb}
\usepackage{amsmath}
\usepackage{mathtools}
\usepackage{geometry}
\usepackage{mathrsfs}
\usepackage{bm}
\usepackage{lineno}
\usepackage{multirow}
\usepackage{booktabs}
\usepackage{listings}
\usepackage{enumerate}
\usepackage{fullpage}
\usepackage{xcolor}
\usepackage{graphicx}
\usepackage{float}
\usepackage{caption}
\usepackage{algorithm}%
\usepackage{algorithmicx}%
\usepackage{algpseudocode}%
\usepackage[colorlinks=true,linkcolor=blue]{hyperref}
\usepackage{url}
\biboptions{numbers,comma,square}
\usepackage{tabularx}

\DeclareMathOperator*{\argmin}{arg\,min}

\journal{}

\begin{document}

\begin{frontmatter}



\title{\textbf{SURGIN}: \textbf{SUR}rogate-guided \textbf{G}enerative \textbf{IN}version for subsurface multiphase flow with quantified uncertainty} 


\author[1,2]{Zhao Feng} 
\author[3]{Bicheng Yan}
\author[4]{Luanxiao Zhao}
\author[1,2]{Xianda Shen}
\author[5]{Renyu Zhao}
\author[6]{Wenhao Wang}
\author[1,2]{Fengshou Zhang\corref{cor}}
\ead{fengshou.zhang@tongji.edu.cn}
\cortext[cor]{Corresponding author}

\affiliation[1]{organization={Department of Geotechnical Engineering, College of Civil Engineering, Tongji University},
            city={Shanghai},
            postcode={200092}, 
            country={China}}
\affiliation[2]{organization={Key Laboratory of Geotechnical and Underground Engineering of Ministry of Education, Tongji University},
            city={Shanghai},
            postcode={200092}, 
            country={China}}
\affiliation[3]{organization={Physical Science and Engineering Division, King Abdullah University of Science and Technology (KAUST)},
            city={Thuwal},
            postcode={23955-6900}, 
            country={Saudi Arabia}}
\affiliation[4]{organization={State Key Laboratory of Marine Geology, Tongji University},
            city={Shanghai},
            postcode={200092}, 
            country={China}}

\affiliation[5]{organization={Tencent SSV},
            city={Beijing},
            postcode={100004}, 
            country={China}}
\affiliation[6]{organization={Tencent SD},
            city={Beijing},
            postcode={100004}, 
            country={China}}

\begin{abstract}
We present a direct inverse modeling method named SURGIN, a SURrogate-guided Generative INversion framework tailed for subsurface multiphase flow data assimilation. Unlike existing inversion methods that require adaptation for each new observational configuration, SURGIN features a zero-shot conditional generation capability, enabling real-time assimilation of unseen monitoring data without task-specific retraining. Specifically, SURGIN synergistically integrates a U-Net enhanced Fourier Neural Operator (U-FNO) surrogate with a score-based generative model (SGM), framing the conditional generation as a surrogate prediction-guidance process in a Bayesian perspective. Instead of directly learning the conditional generation of geological parameters, an unconditional SGM is first pretrained in a self-supervised manner to capture the geological prior, after which posterior sampling is performed by leveraging a differentiable U-FNO surrogate to enable efficient forward evaluations conditioned on unseen observations. Extensive numerical experiments demonstrate SURGIN's capability to decently infer heterogeneous geological fields and predict spatiotemporal flow dynamics with quantified uncertainty across diverse measurement settings. By unifying generative learning with surrogate-guided Bayesian inference, SURGIN establishes a new paradigm for inverse modeling and uncertainty quantification in parametric functional spaces.
\end{abstract}




\end{frontmatter}



\section{Introduction}

Subsurface multiphase flow is a fundamental physical process involved in geo-energy systems such as hydrocarbon extraction~\cite{lyu_role_2021}, geothermal exploitation~\cite{feng_multiphase_2021}, and geological carbon sequestration~\cite{huppert_fluid_2014}. The effective management of these systems necessitates reliable forecasts of subsurface flow dynamics. The predictive accuracy, however, is severely compromised by the uncertainty of reservoir properties, which stems from the intrinsic heterogeneity of subsurface structures and the scarcity of observational data. Inverse modeling, also known as data assimilation, provides a systematic approach to inferring uncertain geological parameters from field observations, thereby improving predictive capability and reducing uncertainty in flow characterization.

Traditional inversion methods are broadly categorized into deterministic and stochastic approaches~\cite{oliver_recent_2011}. Deterministic methods calibrate geological parameters by minimizing the discrepancy between observed reservoir responses and model predictions. Gradient-based methods, which require the computation of sensitivities of the responses with respect to the parameters, include classical algorithms such as the Gauss–Newton method~\cite{tan_fully_nodate, tan_three_1992}, Conjugate Gradient method~\cite{lee_history_1986,makhlouf_general_1993}, Broyden-Flecher-Goldfarb-Shanno method~\cite{zhang_e48_2002,yang_automatic_1988}, and Levenberg–Marquardt method~\cite{zhang_initial_2003,vefring_reservoir_2006}. In contrast, gradient-free methods circumvent the need for explicit derivative information and instead rely on heuristic search strategies. Commonly used techniques include Evolutionary Algorithm~\cite{back1996evolutionary,schulze-riegert_evolutionary_2002}, Neighborhood Algorithm~\cite{sambridge_geophysical_1999,wathelet_improved_2008}, Particle Swarm Optimization~
\cite{luu_parallel_2018,rwechungura_application_2011}, and Pattern Search method~\cite{zhou_pattern-search-based_2012,golmohammadi_pattern-based_2018}, which iteratively explore the parameter space to reduce the mismatch between model predictions and observations. Nevertheless, deterministic approaches provide only point estimates, which precludes full exploration of the posterior parameter space and limits their suitability for uncertainty quantification. Stochastic inversion methods remedy this limitation by treating the unknown geological parameters as random variables and characterizing their posterior distribution according to Bayes’ theorem. A common class of stochastic inversion techniques relies on sampling-based strategies, such as Markov Chain Monte Carlo (MCMC) and its variants~\cite{oliver_markov_1997,efendiev_efficient_2005}, which approximate the posterior distribution through sequentially generated samples. Another widely adopted family is optimization-based ensemble methods~\cite{naevdal_reservoir_2005,elsheikh_parameter_2013}, where an ensemble of realizations is iteratively updated to approximate the Bayesian posterior more efficiently than conventional MCMC. However, traditional inverse modeling methods are computationally intensive, arising from the repeated forward model evaluations and further compounded by the strong nonlinearities, multiphysics couplings, and large spatiotemporal scales in subsurface multiphase flow.

Alternatively, recent advances in machine learning (ML) and deep learning (DL) have enabled the construction of efficient surrogate models that bypass costly forward simulations. Flow dynamics in porous media are governed by parametric partial differential equations (PDEs) that are discretized in space and time. Such discretizations yield grid-based representations that resemble images, making Convolutional Neural Networks (CNNs), originally developed for image processing, well suited to predict fluid flow states (e.g., saturation and pressure) from property fields (e.g., permeability). Notable examples include U-Net based architectures that enhance multiscale feature extraction~\cite{ouyang_co2_2025,tariq2024transunet,mo_deep_2019,yan_physics-constrained_2022}, hybrid models that combine CNNs with sequential networks to capture spatiotemporal dependencies~\cite{feng_encoder-decoder_2024,tang_deep-learning-based_2020,feng_hybrid_2025}, and physics-constrained CNNs designed to enforce consistency with governing flow equations~\cite{qin_fluid_2025,wang_physics-informed_2021,zhang_physics-informed_2023}. Although CNN-based surrogates capture local spatial correlations in grid-based representations, their reliance on fixed discretizations limits generalization, motivating neural operators that directly learn functional mappings between parameter and solution spaces for PDE-based flow modeling~\cite{kovachki_neural_2023,lu_learning_2021}. Within this family, the Fourier Neural Operator~\cite{li_fourier_2021} has emerged as a promising approach, learning infinite-dimensional mappings in Fourier domain and offering decent generalization and predictive accuracy. Wen et al.~\cite{wen_u-fnoenhanced_2022} further extend the vanilla FNO by incorporating a U-Net branch to enhance hierarchical feature learning, which yields improved performance in subsurface multiphase flow modeling. Despite the computational acceleration offered by DL surrogates, their integration into traditional inversion frameworks requires re-running the entire procedure for new observational data, which curtails their generalizability and efficiency in dynamic data assimilation contexts.

Generative modeling, grounded in probabilistic learning and variational inference, provides a promising solution to this challenge. In the context of inverse problems, generative models such as Generative Adversarial Networks (GANs)~\cite{ling_improving_2024,zhan_integrated_2022,laloy_training-image_2018} and Variational Autoencoders (VAEs)~\cite{laloy_inversion_2017,jiang_deep_2021} have been employed to generate geological fields and to assist data assimilation. However, most existing studies have primarily used these models as deterministic dimension-reduction (parameterization) tools, with posterior exploration still dependent on traditional inversion algorithms. As a result, the intrinsic probabilistic capabilities of generative modeling remain largely underexploited. Recently, the score-based generative models (SGMs), have exhibited remarkable performance in high-quality generation and provided a flexible and controllable generating process, finding their applications in a wide range of scientific inverse problems~\cite{du_conditional_2024,dasgupta_conditional_2025,jacobsen_cocogen_2024,huang_diffusionpde_2024,li_learning_2024,gao_bayesian_2024}. SGMs~\cite{song_generative_2019}, which are conceptually equivalent to denoising diffusion probabilistic models (DDPMs)~\cite{ho_denoising_2020}, can be jointly formulated within the stochastic differential equation (SDE)-based framework~\cite{song_score-based_2021}. These models are uniquely characterized by a progressive procedure that transforms data from a simple distribution into a complex one. This is accomplished by gradually perturbing the data with noise and subsequently denoising it through deep neural networks (DNNs). During generation, conditional sampling can be achieved through mechanisms such as classifier guidance~\cite{dhariwal_diffusion_2021}, classifier-free guidance~\cite{ho_classifier-free_2022}, or diffusion posterior sampling~\cite{chung_diffusion_2024}. However, the application of SGMs to subsurface multiphase flow inversion remains relatively unexplored, with only a few recent studies providing preliminary insights in this direction. Fan et al.~\cite{fan_novel_2025} introduce a training-free conditional generative model that leverages a mini-batch-based Monte Carlo estimator for efficient ensemble forecasts in geological carbon storage, without providing posterior geological parameter inference. Zhan et al.~\cite{zhan_toward_2025} propose an integrated latent diffusion model framework for hydrogeological inverse modeling, yet it essentially serves only as a deterministic inverse surrogate. Wang et al.~\cite{wang_generative_2025} employ a conditional diffusion model with classifier-free guidance to directly generate geomodels from sparse observations, but still require a forward simulator to obtain state variables. A key limitation of these studies lies in their reliance on pre-curated condition–state pairs, which undermines zero-shot generalization to unseen observational scenarios. In our most recent work~\cite{feng_generative_2025}, we have taken a leap forward with the development of a Conditional Neural Field Latent Diffusion for geoscience (CoNFiLD-geo) in inverse modeling of geomodels and spatiotemporal state variables within a Bayesian latent space. By leveraging neural fields to implicitly compress high-dimensional parameter and solution spaces, CoNFiLD-geo enables the learning of joint distributions over geomodels and reservoir responses in a shared, mesh-agnostic latent space. CoNFiLD-geo has demonstrated strong performance across various challenging inversion tasks in geological carbon sequestration, producing physics- and observation-consistent results without model retraining. However, co-embedding parameter and solution spaces into a unified latent representation, though convenient, neglects physical causality by obscuring the unidirectional relation imposed by PDEs, wherein parameters unidirectionally determine the state variables.

In this work, we propose a surrogate-guided generative inversion (SURGIN) framework that integrates surrogate modeling with probabilistic generative learning to disentangle the parameter and solution function spaces, using surrogate-predicted solutions to guide exploration of the posterior region in the parameter space. Within SURGIN, an unconditional SGM is first pretrained in a self-supervised manner to capture the prior distribution of geological realizations, while a U-Net enhanced FNO (U-FNO) serves as the forward surrogate to learn the nonlinear functional mapping from geological parameters to state variables. At deployment, observational data can be incorporated through a Bayesian posterior sampling strategy that guides the generative process without requiring retraining. This zero-shot conditional generation capability is enabled by the efficient and differentiable forward surrogate, which gradually steers the learned probability density gradient toward the posterior direction. The posterior state variables can then be readily obtained by inputting the calibrated geomodels into the trained surrogate. We validate the efficacy of SURGIN across a range of subsurface multiphase flow inversion scenarios, including reconstruction from sparse well measurements, recovery of high-fidelity fields from coarsely observations, and restoration of data from corrupted records. The results demonstrate the sound performance of SURGIN in terms of efficiency, generalizability, and robustness, even under high levels of observation sparsity and noise. By enabling real-time data assimilation, uncertainty quantification, and surrogate prediction, SURGIN opens new avenues for both inverse and forward modeling in subsurface flow systems. It bridges the gap between real-world monitoring data and virtual representations, marking a significant step toward realizing digital twins in data-scarce, physics-informed domains, with broader applicability across various subsurface geo-energy systems.

The remainder of this manuscript is structured as follows. Section~\ref{sec2} outlines the problem formulation, and Section~\ref{sec3} introduces the proposed SURGIN framework. Section~\ref{sec4} presents a comprehensive set of numerical experiments to evaluate the inverse modeling capability of SURGIN across various monitoring configurations, using geological carbon sequestration as the demonstration case. Section~\ref{sec5} discusses computational efficiency, uncertainty identification and error propagation within the framework. Finally, Section~\ref{sec6} concludes the manuscript with future perspectives.

\section{Problem statement}\label{sec2}
\subsection{Governing equations of subsurface multiphase flow}
The dynamics of subsurface multiphase flow in porous media are governed by the mass conservation of fluid components~\cite{pruess_tough2_1999},
\begin{equation}\label{governing_eq}
    \frac{\partial}{\partial t}\left(\varphi\sum_\xi \mathbf{S}_\xi \rho_\xi \chi_\xi^\nu\right)
     - \nabla\cdot\left(\mathbf{K}\sum_\xi \chi_\xi^\nu \frac{k_{r,\xi}\rho_\xi}{\eta_\xi}(\nabla\mathbf{P}_\xi-\rho_\xi \mathbf{g}) \right)
     -\sum_\xi\rho_\xi\chi_\xi^\nu q^\nu
     = 0,
\end{equation}
where the first term represents fluid accumulation within the pores, the second term describes the mass flux governed by the extended multiphase Darcy’s law, and the last term accounts for the sources or sinks. Here, $t$ denotes the physical time, the subscript $\xi$ indicates the fluid phase, and the superscript $\nu$ specifies the fluid component. The state variables are the phase saturation $\mathbf{S}_\xi$ and the pressure $\mathbf{P}_\xi$. The formation permeability field is $\mathbf{K}$, the formation porosity is $\varphi$, the phase density is $\rho_\xi$, the phase viscosity is $\eta_\xi$, the mass fraction of component $\nu$ in phase $\xi$ is $\chi_\xi^\nu$, the relative permeability of phase $\xi$ is $k_{r,\xi}$, the gravitational acceleration is $\bm{g}$, and the well volumetric flow rate is $q^\nu$. Eq.~\ref{governing_eq}, together with auxiliary constitutive relationships such as the capillary pressure constraint, relative permeability functions, and fluid thermodynamic equilibrium, defines a closed set of partial differential equations (PDEs) that can be solved numerically to obtain the state variables of saturation and pressure.

\subsection{Formulation of inverse modeling}
Without loss of generality, let $\mathbf{m}\in\mathbb{R}^{N_m}$ signify the geological parameters and $\mathbf{u}\in\mathbb{R}^{N_t\times N_u}$ represent the collection of spatiotemporal state variables, the forward prediction can be expressed as,
\begin{equation}\label{forward_eq}
    \mathbf{u} = \mathcal{F}(\mathbf{m}),
\end{equation}
where $\mathcal{F}:\mathbb{R}^{N_m}\rightarrow\mathbb{R}^{N_t\times N_u}$ denotes the forward operator that satisfies the governing PDEs, $N_t$ is the number of time steps, and $N_m$ and $N_u$ are the dimensions of the geological parameters and state variables, respectively. Eq.~\ref{forward_eq} can be readily solved once $\mathbf{m}$ is fully known a priori. However, the predictive modeling of state variables is fundamentally constrained by incomplete knowledge of formation geological properties, which stems from the intrinsic heterogeneity of subsurface structures. This challenge is further exacerbated by the scarcity of available observations, rendering the forward problem highly uncertain and the corresponding inverse problem severely ill-posed. 

Inverse modeling offers a principled approach to infer the unknown geological parameters by assimilating observational data, thereby enabling the quantification of the associated uncertainty. From a Bayesian probabilistic perspective, inverse modeling seeks to characterize the posterior distribution of geological parameters conditioned on the observed data $\bm{\psi}$,
\begin{equation}\label{bayesian_inverse}
    p(\mathbf{m}|\bm\psi)=\frac{p(\mathbf{m})p(\bm\psi|\mathbf{m})}{p(\bm\psi)},
\end{equation}
where $p(\mathbf{m})$ is prior distribution of geological parameters that encapsulates the initial knowledge of subsurface formation properties, and $p(\bm\psi|\mathbf{m})$ is the likelihood of observed data $\bm\psi$ given geological parameters $\mathbf{m}$. The denominator is a normalizing constant which, however, is generally intractable as it requires integration over the entire parameter space, i.e., $p(\bm\psi)=\int{p(\mathbf{m})p(\bm\psi|\mathbf{m})}d\mathbf{m}$. This complexity can be circumvented by reformulating the expression in terms of the Stein score functions (gradient of the log density)~\cite{chwialkowski2016kernel},
\begin{equation}\label{post_score}
    \nabla_{\mathbf{m}}\log p(\mathbf{m}|\bm\psi)=\nabla_{\mathbf{m}}\log p(\mathbf{m}) + \nabla_{\mathbf{m}}\log p(\bm\psi|\mathbf{m}),
\end{equation}
where the first term $\nabla_{\mathbf{m}}\log p(\mathbf{m})$ corresponds to the score function of the prior distribution of geological parameters, which can be parameterized by an unconditional generative model, while the gradient of the log-likelihood term, $\nabla_{\mathbf{m}}\log p(\bm\psi|\mathbf{m})$, can be evaluated through the forward model $\mathcal{F}$. In the subsequent section, we elaborate on the approximation of these two terms within a surrogate-integrated generative framework, thereby enabling direct sampling from the posterior distribution $p(\mathbf{m}|\bm\psi)$.

\section{The SURGIN framework}\label{sec3}
\subsection{Overview}

\begin{figure}[htb!]
\centering
\includegraphics[width=\textwidth]{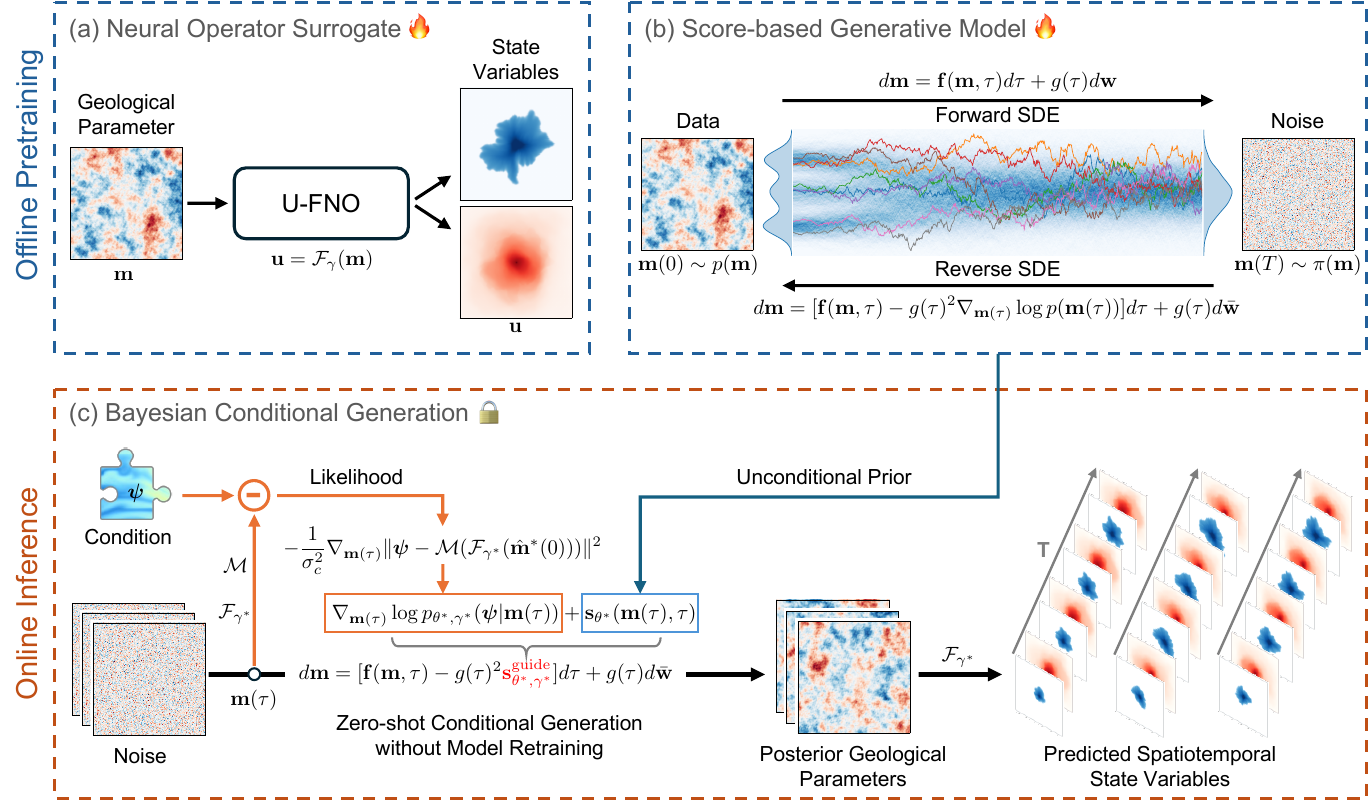}
\caption{Schematic illustration of the SURGIN framework. In the offline phase, (a) a neural operator–based surrogate model and (b) a score-based generative model are pretrained. (c) In the online inference phase, posterior geological parameters are generated via Bayesian conditional generation without retraining, and the corresponding spatiotemporal state variables are predicted by the trained surrogate model.}\label{framework}
\end{figure}

This work presents the SURrogate-guided Generative INversion (SURGIN) model, a probabilistic generative learning framework that synergistically integrates neural operator approximations with generative probabilistic models through a surrogate prediction-guidance mechanism, enabling efficient conditional generation without task-specific retraining. As depicted in Figure~\ref{framework}, SURGIN comprises an offline pretraining stage and an online inference stage.

During offline pretraining, we train two neural networks: (1) a U-Net enhanced Fourier neural operator (U-FNO) surrogate model and (2) a score-based generative model (SGM). The U-FNO surrogate model is trained in a supervised manner to learn the nonlinear functional mapping $\mathcal{F}_\gamma$ from the geological parameter space to the solution space of state variables (Figure~\ref{framework}a). Concurrently, the SGM employs a self-supervised learning scheme to estimate the unconditional score function of the geological parameter prior. This is achieved by initially introducing noise into the data through a predefined forward stochastic differential equation (SDE), and subsequently learning the corresponding reverse SDE by approximating the score function $\nabla_{\mathbf{m}}\log p(\mathbf{m})$ via denoising score matching (Figure~\ref{framework}b).

In the online inference stage, as shown in Figure~\ref{framework}c, SURGIN conditionally generates posterior geological parameters by leveraging the Bayesian formulation and differentiable programming, obviating the need for retraining when adapting to new unseen observations/conditions. This conditional generation process entails solving the reverse SDE using an approximated posterior score function, which is composed of two terms: the unconditional score, learned during pretraining, and the likelihood score, derived from the observational conditions. The latter term guides the reverse SDE's trajectory on the prior probability manifold toward the posterior region by backpropagating the gradient of the discrepancy between surrogate predictions and observations through the entire differentiable framework. By enabling zero-shot conditional generation of observation-consistent geological parameters and their forward predictions of state variables through the trained surrogate model, SURGIN establishes a unified framework for inverse and forward modeling in geoscience, with the capability to quantify uncertainty.

\subsection{U-Net enhanced Fourier neural operator}
While the high-fidelity numerical simulation yields accurate predictions of state variables, the procedure is computationally intensive, and the solver is inherently non-differentiable. Our objective, therefore, is to construct a DL-based surrogate model $\mathcal{F}_\gamma$ that approximates the forward operator $\mathcal{F}$ by learning the functional mapping from the parameter space of geological fields $\mathbf{m}$ to the solution space of reservoir state variables $\mathbf{u}$, i.e., $\mathcal{F}_\gamma:\mathbf{m}\in\mathbb{R}^{N_m}\rightarrow\mathbf{u}\in\mathbb{R}^{N_t\times N_u}$. Specifically, we propose constructing a U-Net enhanced Fourier Neural Operator (U-FNO)~\cite{wen_u-fnoenhanced_2022} that learns an infinite-dimensional space mapping from a finite collection of parameter-state data pairs, enabling efficient predictions of state variables with full differentiability.

\begin{figure}[htb!]
\centering
\includegraphics[width=\textwidth]{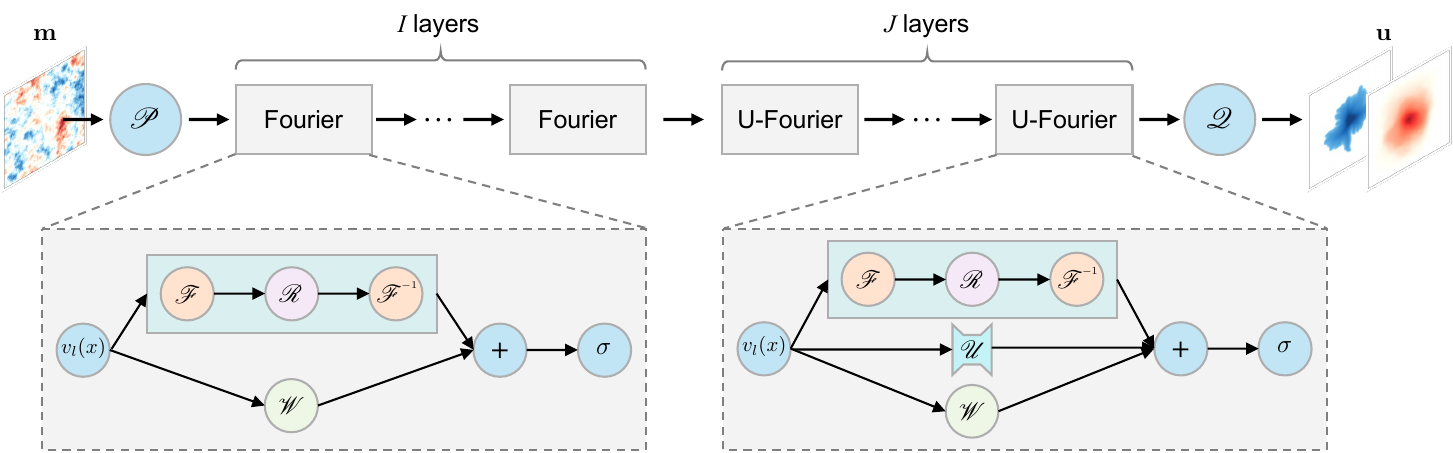}
\caption{Schematic diagram of the U-FNO surrogate model.}\label{ufno_network}
\end{figure}

As illustrated in Figure~\ref{ufno_network}, a U-FNO network $\mathcal{F}_\gamma$ consists of $I$ Fourier layers and $J$ U-Fourier layers, and can be formulated as,
\begin{equation}
    \mathcal{F}_\gamma=\mathscr{Q}\circ\underbrace{\sigma(\mathscr{K}_J+\mathscr{U}_J+\mathscr{W}_J)\circ ... \circ\sigma(\mathscr{K}_1+\mathscr{U}_1+\mathscr{W}_1)}_{\mathrm{U-Fourier\ layers}}\circ\underbrace{\sigma(\mathscr{K}_I+\mathscr{W}_I)\circ... \circ\sigma(\mathscr{K}_1+\mathscr{W}_1)}_{\mathrm{Fourier\ layers}}\circ \mathscr{P},
\end{equation}
where $\mathscr{P}$ and $\mathscr{Q}$ denote the linear lifting and projection operators, respectively. For each intermediate layer, $\mathscr{K}$ represents the kernel integral operator, $\mathscr{U}$ the U-Net convolutional operator, $\mathscr{W}$ a linear transformation, and $\sigma$ a nonlinear activation function. The introduction of the convolutional $\mathscr{U}$ branch essentially enables the network to capture multiscale features, which are ubiquitous in subsurface multiphase flow systems, thereby allowing the network to achieve superior performance compared with the vanilla FNO. The kernel integral operator $\mathscr{K}$ acting on the input function $v_l(x)$ at layer $l$, discretized with respect to $x$, is defined as
\begin{equation}
    \big(\mathscr{K}(v_l)\big)(x) \coloneqq \int\kappa(x,y)v_l(y)dy,
\end{equation}
which is parametrized in Fourier space by using the efficient Fast Fourier Transform (FFT)~\cite{li_fourier_2021},
\begin{equation}
    \big(\mathscr{K}({v}_l)\big)(x) = \mathscr{F}^{-1}\big(\mathscr{R}\cdot\mathscr{F}({v}_l)\big)(x),
\end{equation}
where $\mathscr{F}$ and $\mathscr{F}^{-1}$ represents the Fourier transformation and its inverse, respectively, and $\mathscr{R}$ is the Fourier transform of a periodic function $\kappa$, parameterized by a truncated set of Fourier coefficients.

The U-FNO surrogate model is trained by solving the optimization problem,
\begin{equation}
    \gamma^*=\argmin_\gamma\mathbb{E}_{p(\mathbf{m})}[\| \mathcal{F}_\gamma(\mathbf{m})-\mathcal{F}(\mathbf{m}) \|^2],
\end{equation}
where $\mathbf{m} \sim p(\mathbf{m})$ denotes geological parameter samples drawn from the probability measurement $p(\mathbf{m})$ in the parameter space. In practice, an independent and identically distributed (i.i.d.) dataset $\{(\mathbf{m}^i,\mathbf{u}^i)\}_{i=1}^{N{\mathrm{sample}}}$ is drawn with respect to the probability measurement $p(\mathbf{m})$, with $\mathbf{u}^i = \mathcal{F}(\mathbf{m}^i)$ denoting the corresponding state variables obtained from high-fidelity numerical simulations. Once trained, the surrogate model provides an efficient and differentiable approximation, $\mathbf{u}\simeq\mathcal{F}_{\gamma^*}(\mathbf{m})$, without compromising much predictive accuracy.

\subsection{Score-based unconditional generative model}
An unconditional generative model learns to sample from an unknown distribution given only available samples drawn from such a distribution. Suppose the dataset comprises i.i.d. geological realizations $\{\mathbf{m}^i\in\mathbb{R}^{N_m}\}_{i=1}^{N_{\mathrm{sample}}}$ from the distribution $p(\mathbf{m})$. An SGM, which is parameterized by a score network $\mathbf{s}_\theta:\mathbb{R}^{N_m}\rightarrow\mathbb{R}^{N_m}$, is trained to approximate the Stein score function of the probability density, $\nabla_{\mathbf{m}}\log p(\mathbf{m})$, through score matching~\cite{hyvarinen2005estimation,song_generative_2019}. Once the SGM is trained, sampling from the distribution $p(\mathbf{m})$ can be performed using an iterative process called Langevin dynamics~\cite{parisi_correlation_1981,grenander_representations_1994}. However, one practical limitation of the naive SGM lies in its inability to accurately estimate the score function in low-density regions of the distribution, where the scarcity of samples hinders the evaluation of the score matching objective. A viable remedy is to train the SGM using denoising score matching~\cite{song_generative_2019}, wherein the model learns to approximate the score function of a noise-perturbed distribution. In practice, successive levels of isotropic Gaussian noise are injected into the data until the distribution converges to a tractable form, typically a standard Gaussian that is straightforward to sample from. New samples can then be generated by running the so the so-called annealed Langevin dynamics~\cite{song_generative_2019} in sequence, starting from draws of the tractable Gaussian prior. This procedure is analogous to the forward and reverse Markovian diffusion processes employed in denoising diffusion probabilistic models (DDPMs).~\cite{ho_denoising_2020}. In essence, the iterative procedure of noise injection and removal can be rigorously formulated in a continuous sense through stochastic differential equations (SDEs), which constitute the unified theoretical foundation of both SGM and DDPM.

To this end, we introduce a series of random variables $\{\mathbf{m}(\tau)\}_{\tau=0}^{T}$ indexed by a continuous time parameter $\tau\in[0,T]$, such that $\mathbf{m}(0)\sim p(\mathbf{m})$, where i.i.d geological realizations $\{\mathbf{m}^i\}_{i=1}^{N_{\mathrm{sample}}}$ are drawn, and $\mathbf{m}(T)\sim \pi(\mathbf{m})$, where the distribution admits a tractable form that facilitates efficient sampling. The ordered collection $\{\mathbf{m}(\tau)\}_{\tau=0}^{T}$ corresponds to the solution trajectory of an SDE,
\begin{equation}\label{forward_sde}
    d\mathbf{m} = \mathbf{f}(\mathbf{m},\tau)d\tau + g(\tau)d\mathbf{w},
\end{equation}
where $\mathbf{f}:\mathbb{R}^{N_m}\times\mathbb{R}\rightarrow\mathbb{R}^{N_m}$ is a vector-valued \textit{drift} function, $g:\mathbb{R}\rightarrow\mathbb{R}$ is a scalar-valued \textit{diffusion} function, and $\mathbf{w}$ is the standard Wiener process, also known as Brownian motion. Eq.~\ref{forward_sde} simulates the forward diffusion process of asymptotically adding handcrafted Gaussian noise to $\mathbf{m}(0)$ until the distribution of $\mathbf{m}(\tau)$ converges to a tractable form, typically a standard Gaussian, i.e., $\mathbf{m}(T)\sim\mathcal{N}(\mathbf{0},\mathbf{I})$.

A new sample can be generated from $p(\mathbf{m})$ by starting from the standard Gaussian and reversing the noising process. This reverse denoising process is formulated by solving the SDE backward in time~\cite{anderson_reverse-time_1982},
\begin{equation}\label{backward_sde}
    d\mathbf{m} = [\mathbf{f}(\mathbf{m},\tau)-g(\tau)^2\nabla_{\mathbf{m}(\tau)}\log p(\mathbf{m}(\tau))]d\tau + g(\tau)d\bar{\mathbf{w}},
\end{equation}
where $d\tau$ is an infinitesimal negative timestep running backward from $T$ to $0$, and $d\bar{\mathbf{w}}$ is the standard Wiener process in reverse time. This reverse SDE can be solved once the score of each marginal distribution, $\nabla_{\mathbf{m}(\tau)}\log p(\mathbf{m}(\tau))$, is known for all $\tau$. To this end, a score network $\mathbf{s}_\theta(\mathbf{m}(\tau),\tau)$ is employed to approximate the score function through a parametrized deep neural network such as a U-Net~\cite{nichol_improved_2021} or a diffusion Transformer~\cite{peebles_scalable_2023}. Therefore, the dynamics of Eq.~\ref{backward_sde} are approximated by,
\begin{equation}\label{backward_sde_approx}
    d\mathbf{m} = [\mathbf{f}(\mathbf{m},\tau)-g(\tau)^2\mathbf{s_\theta}(\mathbf{m}(\tau),\tau)]d\tau + g(\tau)d\bar{\mathbf{w}}.
\end{equation}

Following the original formulation of score matching~\cite{hyvarinen2005estimation}, one would ideally train the score network to approximate $\nabla_{\mathbf{m}(\tau)}\log p(\mathbf{m}(\tau))$ at every noise scale, leading to the following loss function,
\begin{equation}
    \mathcal{L}^{\mathrm{SM}}(\theta)=\mathbb{E}_{p(\tau)}\bigg[\lambda(\tau)\mathbb{E}_{p(\mathbf{m}(\tau))}\big[\|\mathbf{s}_\theta(\mathbf{m}(\tau),\tau)-\nabla_{\mathbf{m}(\tau)}\log p(\mathbf{m}(\tau)) \|^2\big]\bigg],
\end{equation}
where $\lambda:[0,T]\rightarrow\mathbb{R}_+$ is a positive weighting function and $p(\tau)\sim\mathcal{U}[0,T]$. This vanilla score matching loss, however, cannot be minimized directly since the score of the marginal distribution,
\begin{equation}
    \nabla_{\mathbf{m}(\tau)}\log p(\mathbf{m}(\tau))=\nabla_{\mathbf{m}(\tau)}\log\int p(\mathbf{m}(\tau)|\mathbf{m}(0))p(\mathbf{m}(0))d\mathbf{m}(0),
\end{equation}
is generally intractable. A practical workaround is to replace the marginal score with the conditional score of the forward transition, which is tractable as long as the drift $\mathbf{f}$ and diffusion $g$ are affine such that the forward SDE defines a Gaussian transition kernel,
\begin{equation}\label{f_trans}
    p(\mathbf{m}(\tau)|\mathbf{m}(0))=\mathcal{N}(\mathbf{m}(\tau);\bm{\mu}(\mathbf{m}(0),\tau),{\sigma}^2(\tau)\mathbf{I}).
\end{equation}

The score network is then trained using the denoising score matching objective~\cite{vincent_connection_2011,song_score-based_2021}, 
\begin{equation}\label{dsm_loss}
    \mathcal{L}^{\mathrm{DSM}}(\theta)=\mathbb{E}_{p(\tau)}\bigg[\lambda(\tau)\mathbb{E}_{p(\mathbf{m}(0))}\mathbb{E}_{p(\mathbf{m}(\tau)|\mathbf{m}(0))}\big[\|\mathbf{s}_\theta(\mathbf{m}(\tau),\tau)-\nabla_{\mathbf{m}(\tau)}\log p(\mathbf{m}(\tau) | \mathbf{m}(0)) \|^2\big]\bigg],
\end{equation}
which can be proved to be equivalent to $\mathcal{L}^{\mathrm{SM}}$ up to a constant, as detailed in~\ref{sm_loss}. In this case, the conditional score admits a closed form for any $\tau$ without explicitly solving the forward SDE,
\begin{equation}\label{score_cond}
    \nabla_{\mathbf{m}(\tau)}\log p(\mathbf{m}(\tau)|\mathbf{m}(0))=-\frac{1}{{\sigma}^2(\tau)}(\mathbf{m}(\tau)-\bm{\mu}(\mathbf{m}(0),\tau)).
\end{equation}

Using the re-parameterization trick~\cite{kingma_auto-encoding_2022},
\begin{equation}
    \mathbf{m}(\tau)=\bm{\mu}(\mathbf{m}(0),\tau)+\sigma(\tau)\bm{\epsilon},
\end{equation}
where $\bm\epsilon\sim\mathcal{N}(\mathbf{0},\mathbf{I})$ is the isotropic Gaussian noise, and substituting this noisy representation back into Eq.~\ref{score_cond}, we further obtain
\begin{equation}\label{noise_obj}
    \nabla_{\mathbf{m}(\tau)}\log p(\mathbf{m}(\tau)|\mathbf{m}(0))=-\frac{\bm{\epsilon}}{\sigma(\tau)},
\end{equation}
which reveals that predicting the score is equivalent to predicting the noise. By re-parameterizing the score network as, 
\begin{equation}
    \mathbf{s}_\theta(\mathbf{m}(\tau),\tau)\coloneqq-\frac{\bm\epsilon_\theta(\bm{m}(\tau),\tau)}{\sigma(\tau)},
\end{equation}
and substituting this together with Eq~\ref{noise_obj} into Eq.~\ref{dsm_loss}, we arrive at the denoising objective in DDPM~\cite{ho_denoising_2020},
\begin{equation}\label{ddpm_loss}
    \mathcal{L}^{\mathrm{DDPM}}(\theta)=\mathbb{E}_{p(\tau)}\bigg[\frac{\lambda(\tau)}{\sigma^2(\tau)}\mathbb{E}_{p(\mathbf{m}(0))}\mathbb{E}_{p(\bm\epsilon)}\big[\|\bm\epsilon_\theta(\bm{m}(\tau),\tau) - \bm\epsilon \|^2\big]\bigg].
\end{equation}

Therefore, the three loss functions are equivalent and can be consistently interpreted within the continuous SDE framework. In this study, we set the weighting function to $\lambda(\tau)=\sigma^2(\tau)$, which simplifies our objective to the unweighted loss function that was empirically found to be effective in~\cite{ho_denoising_2020}. Accordingly, our final training objective is given by,
\begin{equation}\label{ddpm_loss}
    \mathcal{L}^{\mathrm{simple}}(\theta)=\mathbb{E}_{p(\tau),p(\mathbf{m}(0)),p(\bm\epsilon)}\big[\|\bm\epsilon_\theta(\bm{m}(\tau),\tau) - \bm\epsilon \|^2\big].
\end{equation}

\begin{figure}[htb!]
\centering
\includegraphics[width=0.6\textwidth]{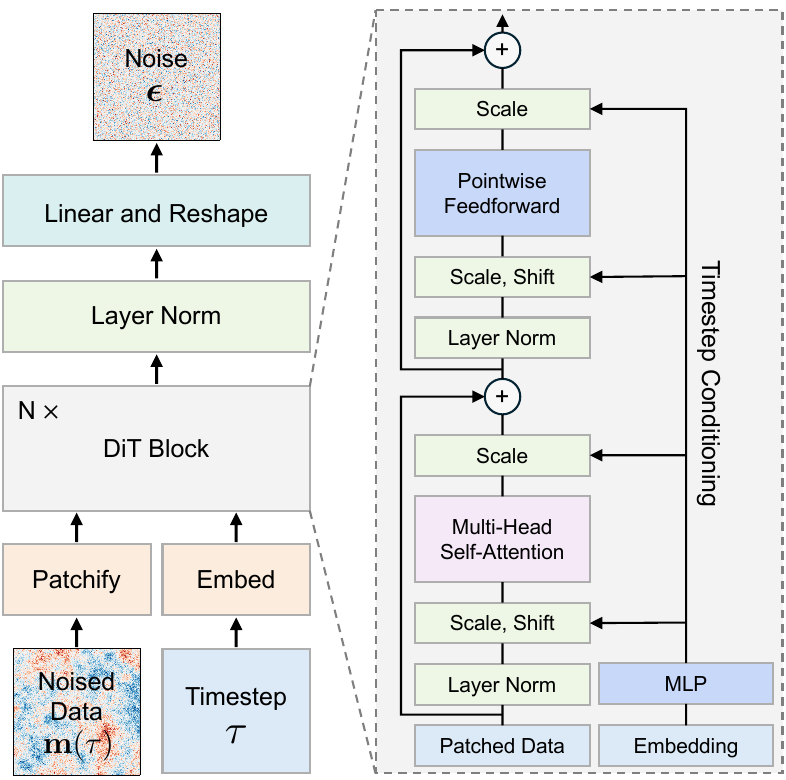}
\caption{Schematic diagram of the DiT score network. The DiT block is a standard Transformer layer augmented with adaptive layer normalization~\cite{perez_film_2018} and scaling to enable timestep conditioning.}\label{dit_network}
\end{figure}

We employ a diffusion Transformer (DiT)~\cite{peebles_scalable_2023} as the score network to predict the added noise over time, owing to its superior performance in computer vision tasks compared with convolution-based U-Nets. The detailed architecture of DiT is illustrated in Figure~\ref{dit_network}. We choose the specific form of $\mathbf{f}$ and $g$ following the variance-preserving (VP) SDE~\cite{song_score-based_2021} formulation as,
\begin{align}
    \mathbf{f}(\mathbf{m},\tau)&=-\frac{1}{2}\beta(\tau)\mathbf{m},\\
    g(\tau)&=\sqrt{\beta(\tau)},
\end{align}
which yields the final forward SDE,
\begin{equation}
    d\mathbf{m} = -\frac{1}{2}\beta(\tau)\mathbf{m}d\tau + \sqrt{\beta(\tau)}d\mathbf{w},
\end{equation}
with the mean and variance of the forward Gaussian transition kernel analytically given by~\cite{sarkka2019applied},
\begin{align}
    \bm{\mu}&=\mathbf{m}(0)\exp{\big(-\frac{1}{2}\int_0^\tau\beta(s)ds\big)},\nonumber\\
    {\sigma}^2&=1-\exp{\big(-\int_0^\tau\beta(s)ds\big)},
\end{align}
where the noise schedule follows the cosine scheme of~\cite{nichol_improved_2021},
\begin{equation}
    \beta(\tau)=\frac{\pi}{1.008T}\tan\bigg(\frac{\tau/T+0.008}{1.008}\cdot\frac{\pi}{2}\bigg).
\end{equation}

The score network is unconditionally pretrained by solving the optimization problem, $\theta^*=\argmin_\theta\mathcal{L}^{\mathrm{simple}}(\theta)$. Once trained, one can use the approximation $\nabla_{\mathbf{m}(\tau)}\log p(\mathbf{m}(\tau))\simeq\mathbf{s}_{\theta^*}(\mathbf{m}(\tau),\tau)$ as a plug-in estimator to replace the score function in the corresponding reverse SDE,
\begin{equation}\label{final_rv_sde}
    d\mathbf{m} = \bigg[-\frac{1}{2}\beta(\tau)\mathbf{m}-\beta(\tau)\nabla_{\mathbf{m}(\tau)}\log p(\mathbf{m}(\tau))\bigg]d\tau + \sqrt{\beta(\tau)}d\bar{\mathbf{w}},
\end{equation}
which can be numerically solved by the Euler–Maruyama scheme, whose updates are analogous to the annealed Langevin dynamics. This procedure enables the unconditional generation of new samples from the Gaussian prior distribution.


    

\subsection{Surrogate guided Bayesian posterior sampling}
In subsurface multiphase flow systems, conditions (i.e., observed data) can be mathematically parameterized by a vector $\bm\psi\in\mathbb{R}^{N_\psi}$. This vector $\bm\psi$ may represent sparse well measurements, low-resolution seismic monitoring data, or any other field observed information pertaining to the spatiotemporal reservior state variables $\mathbf{u}$. In general, the state variables are correlated with the condition vector through the relation,
\begin{equation}\label{M_op}
    \bm\psi=\mathcal{M}(\mathbf{u})+\bm\epsilon_c,
\end{equation}
where $\mathcal{M}:\mathbb{R}^{N_t\times N_u}\rightarrow\mathbb{R}^{N_\psi}$ is a nonlinear functional mapping from the state variables $\mathbf{u}$ to their associated condition vector $\bm\psi$, $\bm\epsilon_c$ denotes the measurement noise arising from monitoring uncertainty, typically modeled as a zero-mean Gaussian with variance $\sigma^2_c$. The mapping $\mathcal{M}$ is essentially a spatiotemporal masking operator, which formalizes the fact that only a subset of the full state variables $\mathbf{u}$ can be observed in practice. By replacing the true state variables $\mathbf{u}$ with their surrogate approximations, $\mathbf{\hat u}=\mathcal{F}_{\gamma^*}(\mathbf{m})$, Eq.~\ref{M_op} can be reformulated as,
\begin{equation}\label{M_op2}
    \bm\psi\simeq\mathcal{M}(\mathcal{F}_{\gamma^*}(\mathbf{m}))+\bm\epsilon_c,
\end{equation}
where $\mathcal{F}_{\gamma^*}$ denotes the trained U-FNO surrogate model that maps the geological parameter $\mathbf{m}$ to the corresponding state variables $\mathbf{u}$. In the case of Gaussian measurement noise, the relation in Eq.~\ref{M_op2} further implies the following conditional probability,
\begin{equation}\label{M_op_cond_prob}
    p(\bm\psi|\mathbf{m})\sim\mathcal{N}(\bm\psi;\mathcal{M}(\mathcal{F}_{\gamma^*}(\mathbf{m})),\sigma_c^2\mathbf{I}).
\end{equation}

The conditional generation of the posterior geological field $\mathbf{m}$ given observation $\bm\psi$ corresponds to approximating the score function of the posterior distribution, $\nabla_{\mathbf{m}(\tau)}\log p(\mathbf{m}(\tau)|\bm\psi)$, at every noise level $\tau$. As indicated in Eq~\ref{post_score}, this posterior score can be reformulated via Baye's rule,
\begin{equation}\label{score_posterior_final}
    \nabla_{\mathbf{m}(\tau)}\log p(\mathbf{m}(\tau)|\bm\psi)=\nabla_{\mathbf{m}(\tau)}\log p(\mathbf{m}(\tau))+\nabla_{\mathbf{m}(\tau)}\log p(\bm\psi|\mathbf{m}(\tau)),
\end{equation}
where $\nabla_{\mathbf{m}(\tau)}\log p(\mathbf{m}(\tau))$ has been learned by the unconditional SGM during the pretraining stage, i.e., $\nabla_{\mathbf{m}(\tau)}\log p(\mathbf{m}(\tau))\simeq\mathbf{s}_{\theta^*}(\mathbf{m}(\tau),\tau)$, while the likelihood score $\nabla_{\mathbf{m}(\tau)}\log p(\bm\psi|\mathbf{m}(\tau))$ guides the generated $\mathbf{m}(\tau)$ toward consistency with $\bm{\psi}$ and requires approximation on the probability manifold.

We first develop the likelihood term as,
\begin{align}\label{factorize}
    p(\bm\psi|\mathbf{m}(\tau))&=\int p(\bm\psi|\mathbf{m}(0),\mathbf{m}(\tau))p(\mathbf{m}(0)|\mathbf{m}(\tau))d\mathbf{m}(0),\nonumber\\
    &=\int p(\bm\psi|\mathbf{m}(0))p(\mathbf{m}(0)|\mathbf{m}(\tau))d\mathbf{m}(0),\nonumber\\
    &=\mathbb{E}_{\mathbf{m}(0)\sim p(\mathbf{m}(0)|\mathbf{m}(\tau))}[ p(\bm\psi|\mathbf{m}(0))],
\end{align}
where the second equality holds based on the fact that $\bm\psi$ and $\mathbf{m}(\tau)$ are conditionally independent on $\mathbf{m}(0)$. Although the forward transition (Eq.~\ref{f_trans}) is predefined as a Gaussian kernel, the reversed transition, $p(\mathbf{m}(0)|\mathbf{m}(\tau))$, is generally intractable. Nevertheless, a distinctive property of the VP SDE is that this conditional distribution $p(\mathbf{m}(0)|\mathbf{m}(\tau))$ admits a unique posterior mean, which can be analytically obtained through the Tweedie’s formula~\cite{efron_tweedies_2011,kim_noise2score_2021},
\begin{equation}\label{pst_mean}
    \hat{\mathbf{m}}(0)\coloneqq\mathbb{E}[\mathbf{m}(0)|\mathbf{m}(\tau)]=\frac{1}{\sqrt{\bar{\alpha}(\tau)}}\bigg(\mathbf{m}(\tau)+(1-\bar{\alpha}(\tau))\nabla_{\mathbf{m}(\tau)}\log p(\mathbf{m}(\tau))\bigg),
\end{equation}
where $\bar{\alpha}(\tau)=\exp{\big(-\int_0^\tau\beta(s)ds\big)}$ is the cumulative noise attenuation factor in the continuous setting. By substituting the score estimate $\mathbf{s}_{\theta^*}(\mathbf{m}(\tau),\tau)$ for $\nabla_{\mathbf{m}(\tau)}\log p(\mathbf{m}(\tau))$ in Eq.~\ref{pst_mean}, the posterior mean of $p(\mathbf{m}(0)\mid \mathbf{m}(\tau))$ can be approximated by the pretrained unconditional SGM,
\begin{equation}
    \hat{\mathbf{m}}^*(0)\simeq\frac{1}{\sqrt{\bar{\alpha}(\tau)}}\bigg(\mathbf{m}(\tau)+(1-\bar{\alpha}(\tau))\mathbf{s}_{\theta^*}(\mathbf{m}(\tau),\tau)\bigg).
\end{equation}

Since the posterior mean $\hat{\mathbf{m}}(0)$ can be computed at the intermediate steps, the likelihood in Eq.~\ref{factorize} can be approximated by using the Jensen's inequality~\cite{chung_diffusion_2024},
\begin{equation}
    p(\bm\psi|\mathbf{m}(\tau))\simeq p(\bm\psi|\hat{\mathbf{m}}(0)),
\end{equation}
where $\hat{\mathbf{m}}(0)\coloneqq\mathbb{E}[\mathbf{m}(0)|\mathbf{m}(\tau)]=\mathbb{E}_{\mathbf{m}(0)\sim p(\mathbf{m}(0)|\mathbf{m}(\tau))}[ p(\mathbf{m}(0))]$, and the approximation error is theoretically bounded by the Jensen gap~\cite{simic_global_2008,gao_bounds_2020}.

Now, based on the conditional probability outlined in Eq.~\ref{M_op_cond_prob}, we can explicitly evaluate the likelihood score as,
\begin{equation}
    \nabla_{\mathbf{m}(\tau)}\log p(\bm\psi|\mathbf{m}(\tau))\simeq
    -\frac{1}{\sigma_c^2}\nabla_{\mathbf{m}(\tau)}\| \bm\psi -  \mathcal{M}(\mathcal{F}_{\gamma^*}(\hat{\mathbf{m}}^*(0)))\|^2,
\end{equation}
which is computationally tractable via the chain rule,
\begin{align}
    \nabla_{\mathbf{m}(\tau)}\log p(\bm\psi|\mathbf{m}(\tau))&\simeq\nabla_{\mathbf{m}(\tau)}\log p_{\theta^*,\gamma^*}(\bm\psi|\mathbf{m}(\tau))\nonumber\\
    &=-\frac{2}{\sigma_c^2}\bigg(\bm\psi -  \mathcal{M}(\mathcal{F}_{\gamma^*}(\hat{\mathbf{m}}^*(0)))\bigg)\frac{\partial \mathcal{M}(\mathcal{F}_{\gamma^*})}{\partial \mathcal{F}_{\gamma^*}}\frac{\partial \mathcal{F}_{\gamma^*}(\hat{\mathbf{m}}^*;\gamma^*)}{\partial \hat{\mathbf{m}}^*}\frac{\partial \hat{\mathbf{m}}^*(\mathbf{m}(\tau);\theta^*)}{\partial \mathbf{m}(\tau)},
\end{align}
by leveraging the automatic differentiation (AD) capability as the entire framework is implemented in a fully differentiable manner in PyTorch~\cite{paszke_pytorch_2019}.

Finally, the gradient of log posterior as shown by Eq.~\ref{score_posterior_final} can be computed as,
\begin{align}\label{s_guide}
    \nabla_{\mathbf{m}(\tau)}\log p(\mathbf{m}(\tau)|\bm\psi)&\simeq\nabla_{\mathbf{m}(\tau)}\log p_{\theta^*}(\mathbf{m}(\tau)) + \nabla_{\mathbf{m}(\tau)}\log p_{\theta^*,\gamma^*}(\bm\psi|\mathbf{m}(\tau))\nonumber\\
    &=\mathbf{s}_{\theta^*}(\mathbf{m}(\tau),\tau)+\nabla_{\mathbf{m}(\tau)}\log p_{\theta^*,\gamma^*}(\bm\psi|\mathbf{m}(\tau))\nonumber\\
    &=\mathbf{s}_{\theta^*,\gamma^*}^{\mathrm{guide}}(\bm\psi,\mathbf{m}(\tau),\tau;\theta^*,\gamma^*),
\end{align}
where $\mathbf{s}_{\theta^*,\gamma^*}^{\mathrm{guide}}$ is the surrogate guided conditional score function. Note that $\mathbf{s}_{\theta^*,\gamma^*}^{\mathrm{guide}}$ is composed simply by adding the pretrained unconditional score to the likelihood term evaluated through the forward model, the latter being efficiently approximated by the U-FNO surrogate. Therefore, at inference time, the posterior geological parameters $\mathbf{m}|\bm\psi$ given conditions $\bm\psi$ can be generated in a plug-and-play manner by solving the reverse SDE without model retraining,
\begin{equation}\label{guided_sample}
    d\mathbf{m} = \bigg[-\frac{1}{2}\beta(\tau)\mathbf{m}-\beta(\tau)\mathbf{s}_{\theta^*,\gamma^*}^{\mathrm{guide}}(\bm\psi,\mathbf{m}(\tau),\tau;\theta^*,\gamma^*)\bigg]d\tau + \sqrt{\beta(\tau)}d\bar{\mathbf{w}}.
\end{equation}

\section{Numerical experiments}\label{sec4}
In this section, we validate SURGIN on representative subsurface multiphase flow scenarios, spanning inversion from sparse well measurements, low-resolution observations, and corrupted records, demonstrating its robustness and efficiency in conditional generation for inverse modeling with quantified uncertainty. The training and inference protocols are presented in~\ref{hyperparam}, while the hyperparameters of U-FNO and DiT are provided in~\ref{ufno} and~\ref{dit}, respectively. The focus of these numerical experiments, rather than on complicating subsurface flow configurations, is to showcase SURGIN’s ability to adaptively assimilate diverse observational information $\bm\psi$ without task-specific retraining. We remark, however, that the proposed framework is generalizable and can be readily applied to a broad spectrum of subsurface multiphase flow inverse problems with minimal architectural modifications. 

\subsection{Case 1: horizontal model}
\subsubsection{Numerical model setup}
We first consider the injection of supercritical $\mathrm{CO_2}$ into a 2-dimensional horizontal porous reservoir with a domain of $5~\mathrm{km}\times5~\mathrm{km}\times50~\mathrm{m}$, discretized into $64\times64\times1$ uniform grids. An injection well is located at the geometric center of the square domain, operating at a constant rate of 32 kg/s, which corresponds to a $\mathrm{CO_2}$ sequestration capacity of approximately 1 Mt per year. The simulation spans 20 years, with outputs collected at 1 month, 6 months, and at 1, 2, 3, 5, 7, 10, 15, and 20 years, resulting in 10 time steps in total. The four lateral boundaries are prescribed as open-flow boundaries, while the top and bottom boundaries are treated as impermeable. The horizontal reservoir is characterized by heterogeneous permeability fields, designed to mimic the uncertain distribution of geological parameters. The heterogeneity is modeled as a Gaussian random field parameterized by a covariance kernel with an isotropic correlation length of 1250~m~\cite{muller_gstools_2022}. The field is assigned a log-normal distribution with mean $\ln(100)$ (geometric mean 100~mD) and standard deviation 1.0. The initial pressure and temperature are set to 20 MPa and $70^{\circ}\mathrm{C}$, respectively, and the model is assumed to be isothermal. The porosity is set to a uniform value of 0.2 across the entire domain. The Van Genuchten capillary pressure model and the Corey relative permeability model are employed, with parameters adopted from~\cite{tang_deep-learning-based_2022}. The forward simulation is solved by the high fidelity numerical solver TOUGH3/ECO2N~\cite{jung_tough3_2017}. A total of 2,000 input–output pairs are collected for pretraining, and an additional 200 pairs are reserved as the testing dataset for inverse modeling.

Our aim is to infer the uncertain geological parameters $\mathbf{m}$ (i.e., permeability fields) from limited observations of the state variables $\mathbf{u}$ (i.e., $\mathrm{CO_2}$ saturation and pressure). There are circumstances, however, where direct measurements of geological parameters are available, for example from well logs or rock sample tests~\cite{ahmed_permeability_1991}. In this setting, the guided conditional score in Eq.~\ref{s_guide} is modified to incorporate the conditional likelihood scores of direct parameter observations, which does not require evaluation through the forward operator $\mathcal{F}$. The corresponding guided conditional score function is expressed as,
\begin{align}\label{s_guide_2}
    \mathbf{s}_{\theta^*,\gamma^*}^{\mathrm{guide}}(\bm\psi,\mathbf{m}(\tau),\tau;\theta^*,\gamma^*)=\mathbf{s}_{\theta^*}(\mathbf{m}(\tau),\tau)+\nabla_{\mathbf{m}(\tau)}\log p_{\theta^*,\gamma^*}(\bm\psi|\mathbf{m}(\tau))+\nabla_{\mathbf{m}(\tau)}\log p_{\theta^*}(\bm\psi|\mathbf{m}(\tau)),
\end{align}
where the second term can be omitted in the absence of observations of state variable, the last term corresponds to direct observations of geological parameters, approximated by,
\begin{equation}
    \nabla_{\mathbf{m}(\tau)}\log p_{\theta^*}(\bm\psi|\mathbf{m}(\tau))\simeq
    -\frac{1}{\sigma_c^2}\nabla_{\mathbf{m}(\tau)}\| \bm\psi -  \mathcal{M}(\hat{\mathbf{m}}^*(0))\|^2.
\end{equation}

In case 1, we consider two challenging observation scenarios for inverse modeling: (1) sparse well measurements and (2) low-fidelity observational data. For each scenario, we conditionally generate an ensemble of 20 geological realizations and employ the surrogate model to predict the corresponding state variables. The observational information $\bm\psi$, that is, the conditions used to guide the generation process, is drawn entirely from the test dataset and has never been seen by the model during training. The associated uncertainty is quantified by the standard deviation of the ensemble. To rigorously evaluate the performance of inverse and forward modeling, we adopt the Kullback–Leibler (KL) divergence and the Structural Similarity Index Measure (SSIM) as evaluation metrics, expressed as,
\begin{equation}
    D_{\mathrm{KL}}(p_{\mathrm{ref}}(\mathbf{m})\|p_{\mathrm{gen}}(\mathbf{m}))=\int p_{\mathrm{ref}}(\mathbf{m})\ln{\frac{p_{\mathrm{ref}}(\mathbf{m})}{p_{\mathrm{gen}}(\mathbf{m})}}d\mathbf{m},
\end{equation}
\begin{equation}
    \mathrm{SSIM}(\mathbf{u},\hat{\mathbf{u}})=\frac{1}{M}\sum_{m}^{M}\frac{(2\mu_{\mathbf{u},m}\mu_{\hat{\mathbf{u}},m}+K_1^2)(2\sigma_{\mathbf{u}\hat{\mathbf{u}},m}+K_2^2)}{(\mu_{\mathbf{u},m}^2+\mu_{\hat{\mathbf{u}},m}^2+K_1^2)(\sigma_{\mathbf{u},m}^2+\sigma_{\hat{\mathbf{u}},m}^2+K_2^2)},
\end{equation}
where $p_{\mathrm{ref}}(\mathbf{m})$ and $p_{\mathrm{gen}}(\mathbf{m})$ are the distribution of the reference and inferred geological parameters, $\mathbf{u}$ and $\hat{\mathbf{u}}$ are the reference and predicted state variables, $M$ is the number of local windows used to evaluate the snapshots/images, $\mu$ and $\sigma$ are the mean and standard deviation, and $K_1$ and $K_2$ are small constants introduced to prevent division by zero. KL divergence quantifies the discrepancy between two probability distributions, with smaller non-negative values indicating closer agreement. SSIM assesses the resemblance between two images, where values approaching 1 indicate higher similarity.

\subsubsection{Reconstruction based on sparse well measurements}

\begin{figure}[htb!]
\centering
\includegraphics[width=0.95\textwidth]{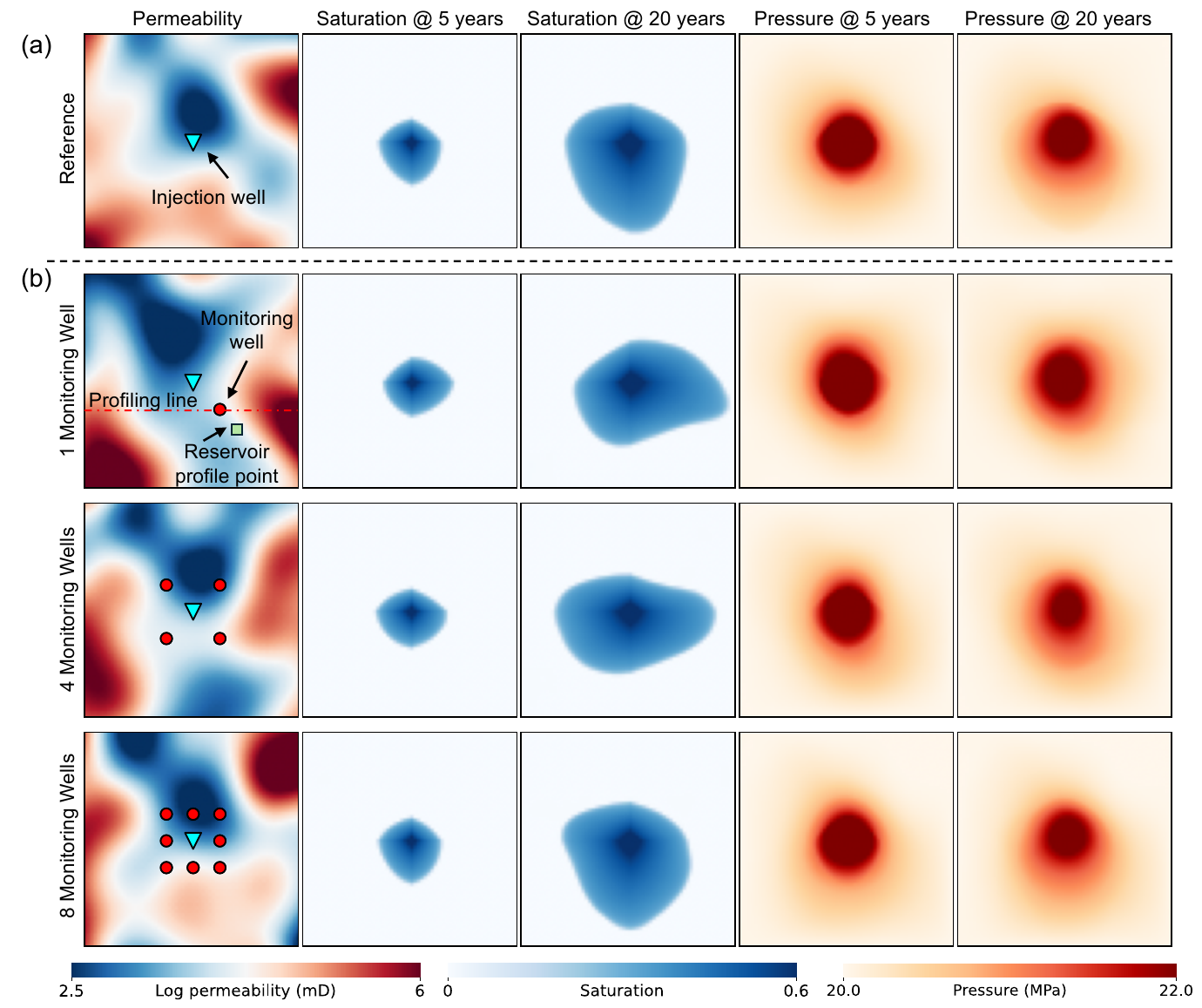}
\caption{(a) Reference permeability fields and state variables ($\mathrm{CO_2}$ saturation and pressure) at 5 and 20 years. (b) Generated permeability fields and predicted state variables conditioned on observations of state variables from 1, 4, and 8 monitoring wells.}\label{fig_case1_sp1}
\end{figure}

\begin{figure}[htb!]
\centering
\includegraphics[width=0.95\textwidth]{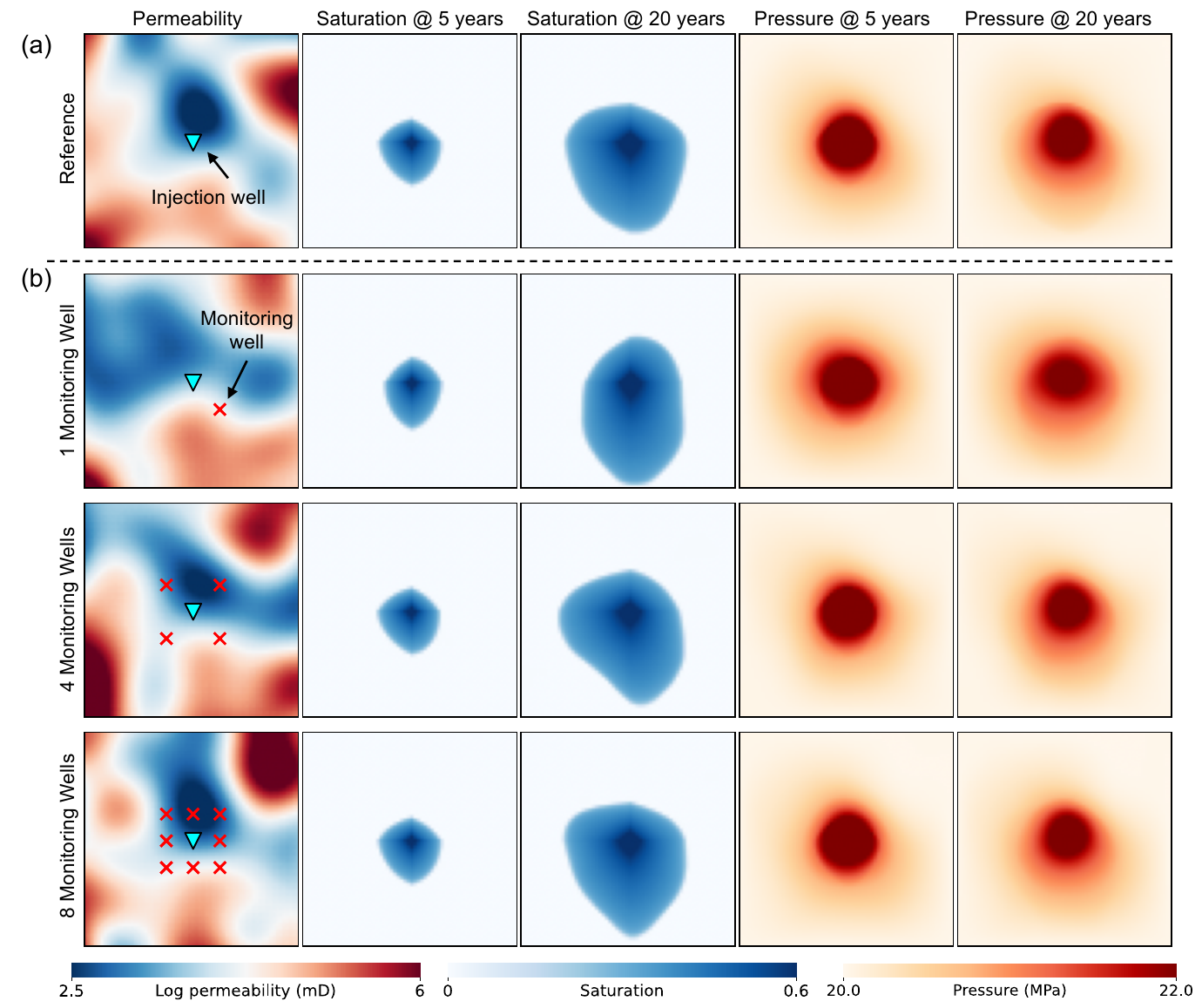}
\caption{(a) Reference permeability fields and state variables ($\mathrm{CO_2}$ saturation and pressure) at 5 and 20 years. (b) Generated permeability fields and predicted state variables conditioned on observations of both state variables and permeability values from 1, 4, and 8 monitoring wells.}\label{fig_case1_sp2}
\end{figure}

\begin{figure}[htb!]
\centering
\includegraphics[width=0.85\textwidth]{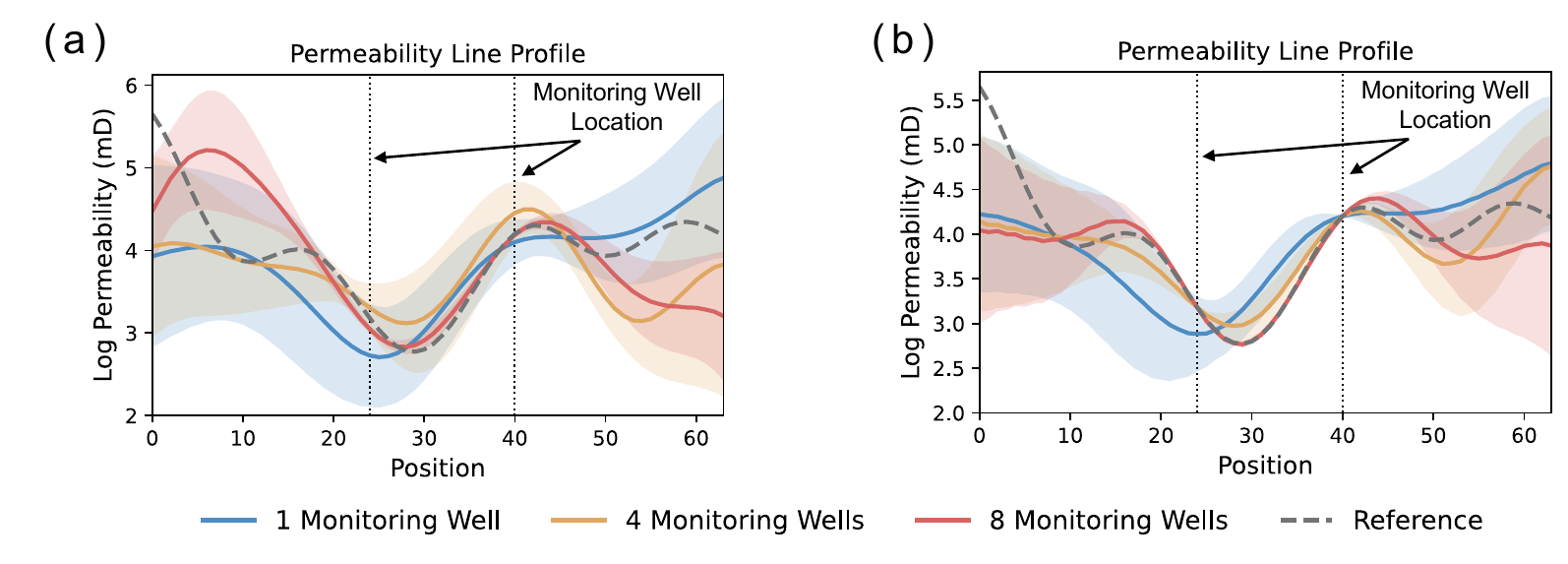}
\caption{Permeability profiles along the profiling line, with dotted lines indicating monitoring well locations. Shaded areas represent uncertainty (standard deviation). (a) Conditioned on observations of state variables. (b) Conditioned on observations of both state variables and permeability values.}\label{fig_case1_sp3}
\end{figure}

\begin{figure}[htb!]
\centering
\includegraphics[width=0.8\textwidth]{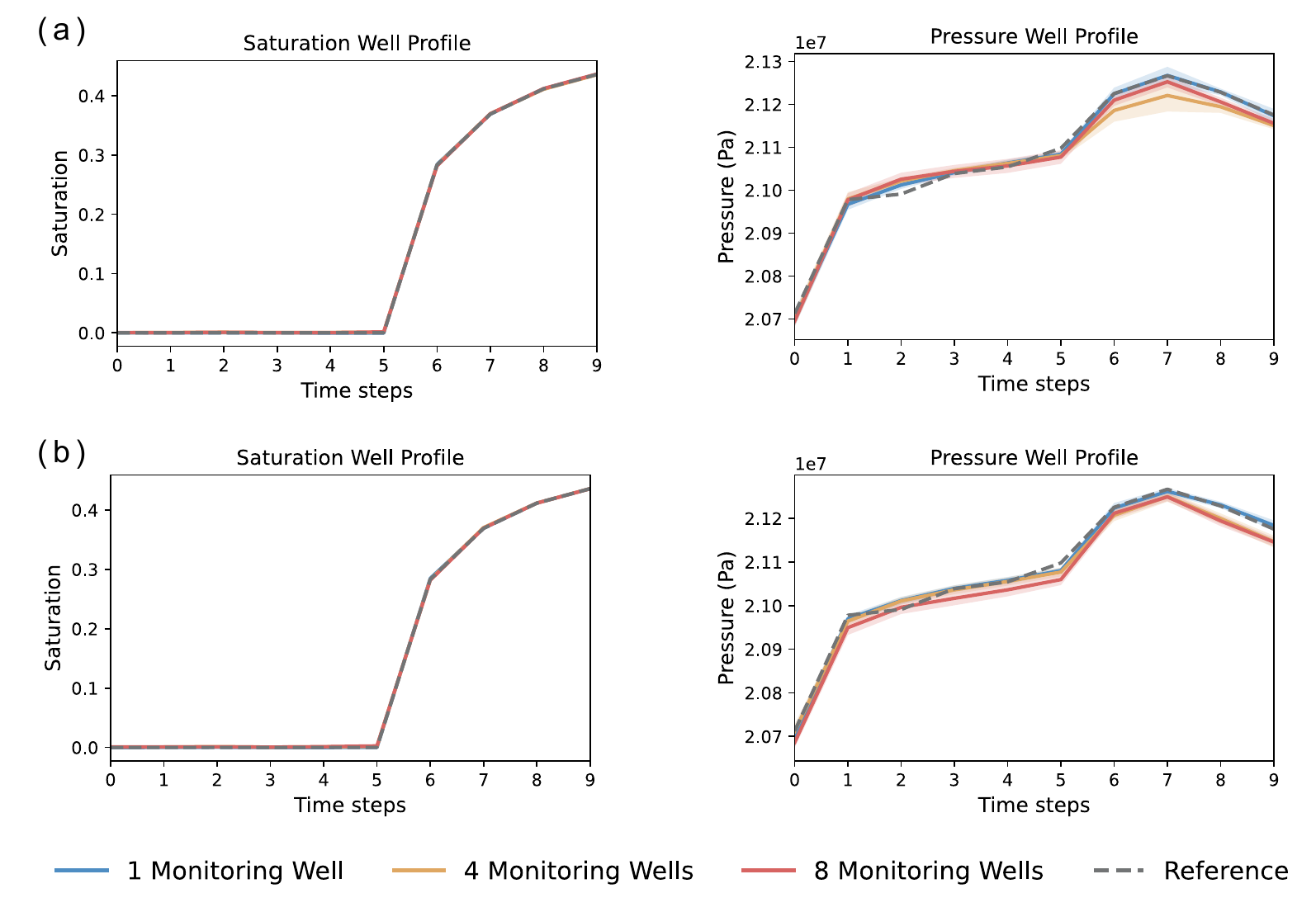}
\caption{Temporal profiles of state variables at the monitoring well location, with shaded areas denoting uncertainty. (a) Conditioned on observations of state variables. (b) Conditioned on observations of both state variables and permeability values.}\label{fig_case1_sp4}
\end{figure}

\begin{figure}[htb!]
\centering
\includegraphics[width=0.8\textwidth]{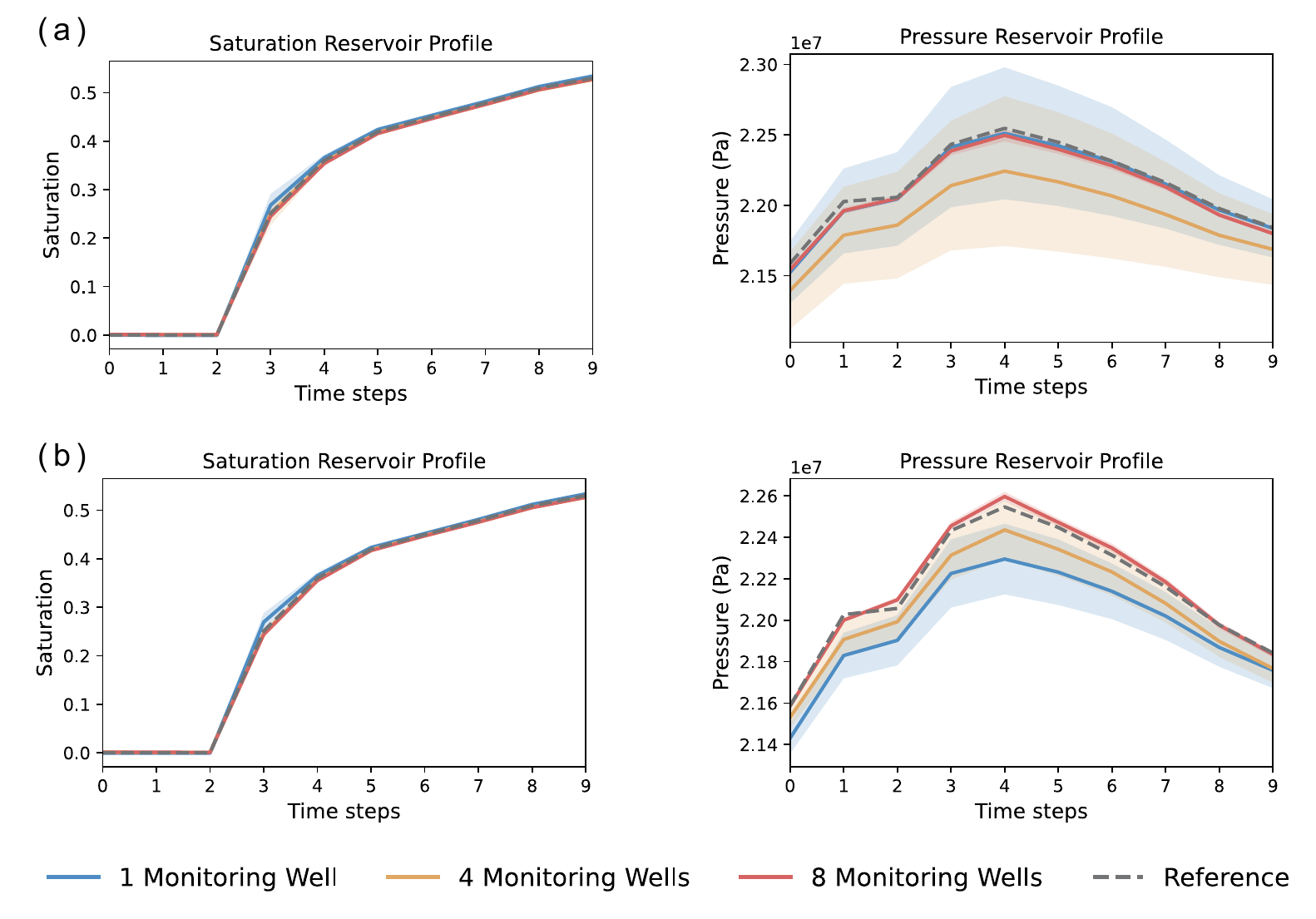}
\caption{Temporal profiles of state variables at the reservoir profiling location, with shaded areas denoting uncertainty. (a) Conditioned on observations of state variables. (b) Conditioned on observations of both state variables and permeability values.}\label{fig_case1_sp5}
\end{figure}

\begin{figure}[htb!]
\centering
\includegraphics[width=0.8\textwidth]{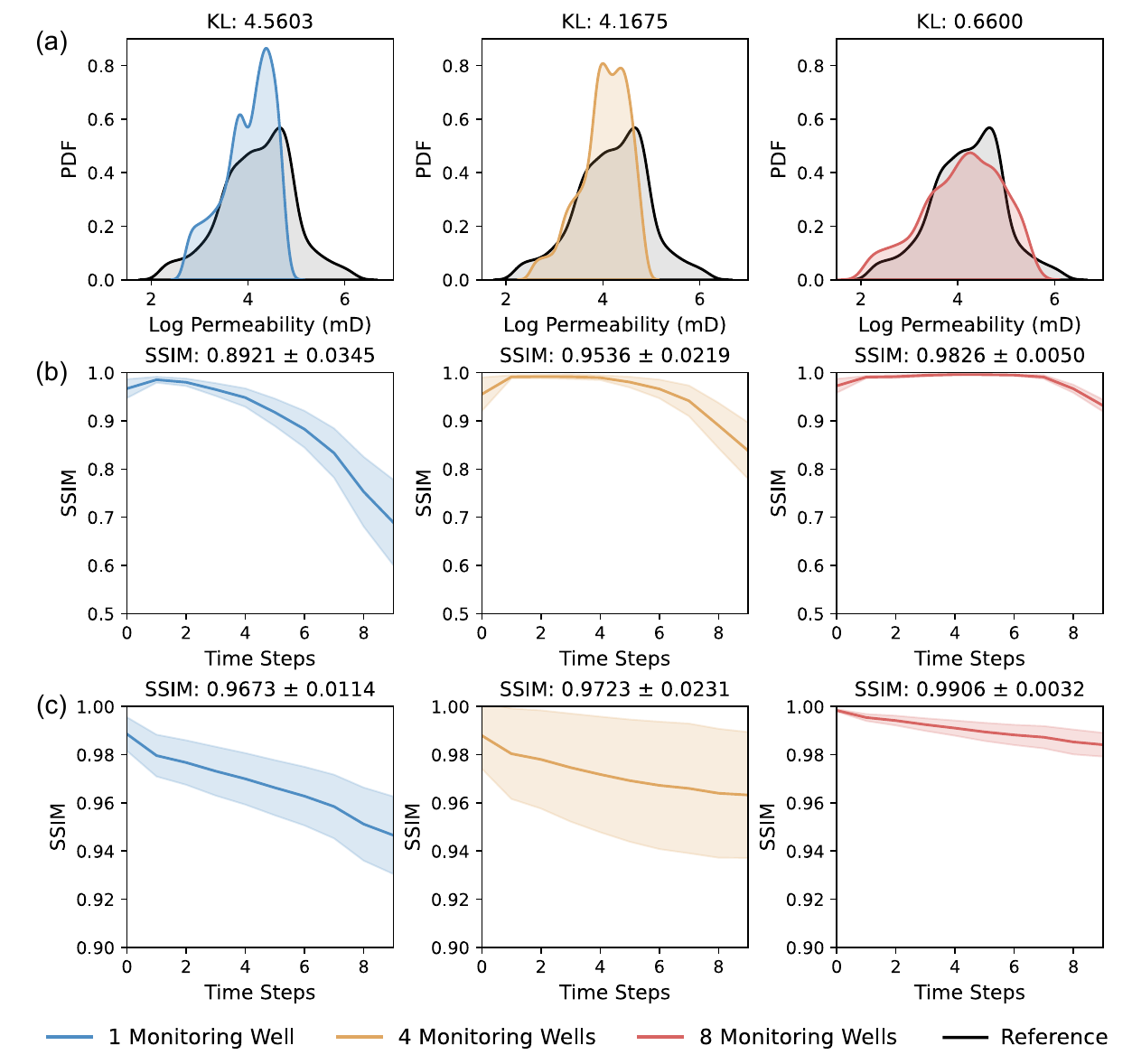}
\caption{Quantitative metrics evaluated over all generated ensemble samples conditioned on observations of state variables. (a) Probability density function (PDF) of the generated permeability fields compared with the reference, with the corresponding KL divergence. (b) SSIM of saturation between generated and reference trajectories, with shaded areas denoting the standard deviation. (c) SSIM of pressure between generated and reference trajectories.}\label{fig_case1_sp6}
\end{figure}

\begin{figure}[htb!]
\centering
\includegraphics[width=0.8\textwidth]{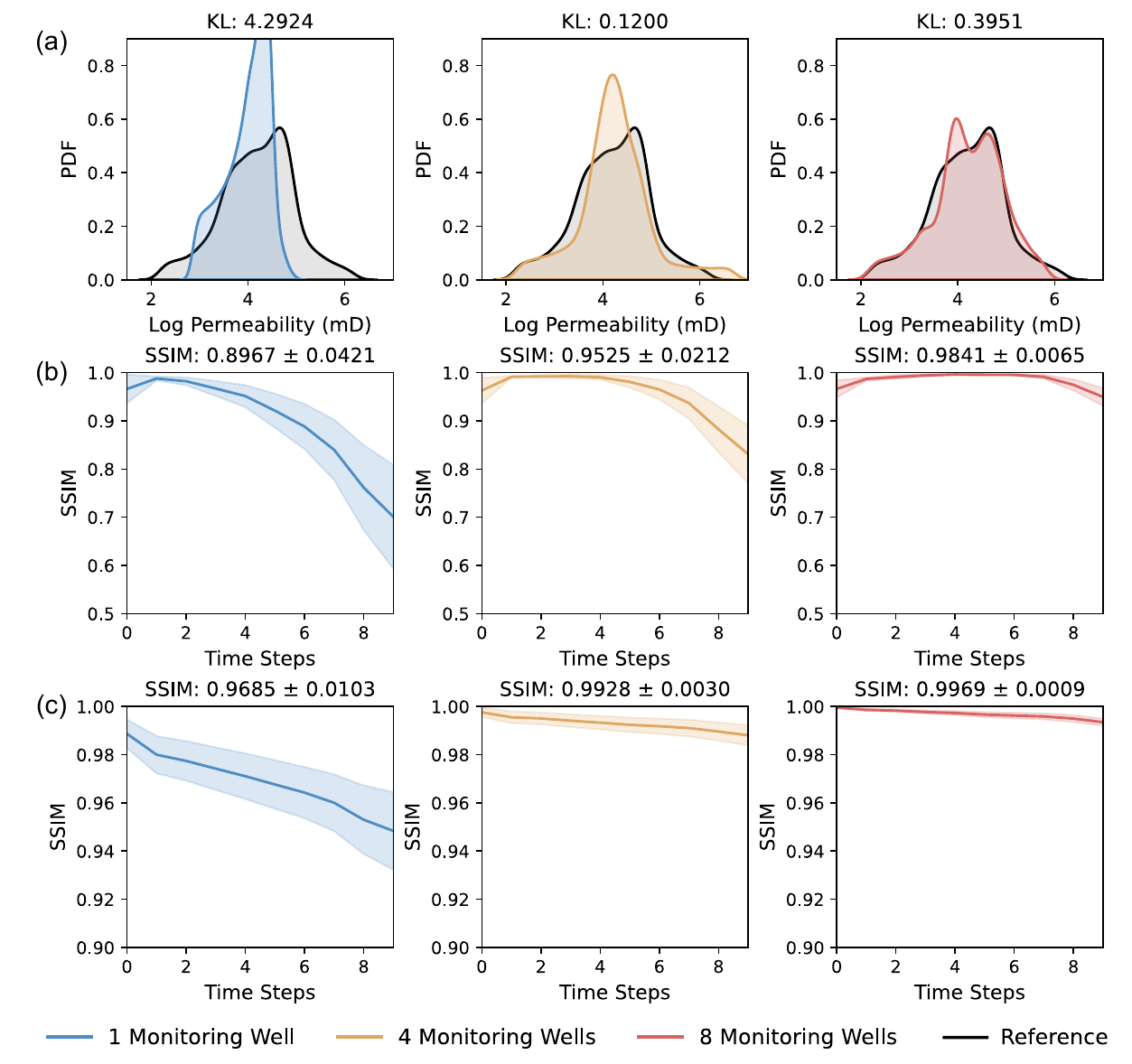}
\caption{Quantitative metrics evaluated over all generated ensemble samples conditioned on observations of both state variables and permeability values. (a) PDF of the generated permeability fields compared with the reference, with the corresponding KL divergence. (b) SSIM of saturation between generated and reference trajectories, with shaded areas denoting the standard deviation. (c) SSIM of pressure between generated and reference trajectories.}\label{fig_case1_sp7}
\end{figure}

In subsurface multiphase flow systems, drilling wells provides intrusive measurements of flow states and reservoir properties, offering critical but spatially limited insights into the underground system~\cite{massarweh_co2_2024}. Since well drilling and logging are costly and technically constrained, the number of wells is typically sparse relative to the vast spatial extent of the reservoir. This sparsity leads to incomplete characterization of the subsurface, posing significant challenges for data assimilation and inverse modeling. To address this challenge, we demonstrate the efficacy of our proposed method in generating observation-consistent geological parameters and state variables from sparse well measurements, providing robust solutions to such inverse problems. Figure~\ref{fig_case1_sp1} and~\ref{fig_case1_sp2} display the generated permeability fields and the predicted saturation and pressure snapshots conditioned on 1, 4, and 8 monitoring wells (Panel (b)), shown alongside the ground-truth reference for comparison (Panel (a)). The results in Figure~\ref{fig_case1_sp1} are conditioned solely on saturation and pressure data, whereas those in Figure~\ref{fig_case1_sp2} additionally incorporate direct measurements of permeability values. As the number of monitoring wells increases, the inferred permeability fields more closely resemble the reference, and the predicted saturation and pressure dynamics become more consistent with the ground truth, in accordance with the Bayesian perspective. Incorporating direct measurements of reservoir properties helps to better constrain both the geological parameters and the state variables, as further substantiated in Figure~\ref{fig_case1_sp3},~\ref{fig_case1_sp4}, and ~\ref{fig_case1_sp5}. The uncertainty of the geological parameters decreases markedly in the vicinity of well locations where permeability values are probed (Figure~\ref{fig_case1_sp3}b). Likewise, the uncertainty of the state variables across the temporal domain is consistently lower at the well locations (Figure~\ref{fig_case1_sp4}) than at other reservoir locations (Figure~\ref{fig_case1_sp5}). This trend is expected, as conditioning information naturally constrains the posterior space, whereas uncertainty grows farther from the probed data due to the inherent ill-posedness and non-uniqueness of the inverse problem. Remarkably, our proposed method faithfully captures this behavior and provides a principled quantification of the associated uncertainty without the need for model retraining, offering a systematic understanding of the role of observational strength in determining inversion outcomes. Furthermore, the overall performance, evaluated by KL divergence and SSIM (Figure~\ref{fig_case1_sp6} and~\ref{fig_case1_sp7}), demonstrates the robustness of our method in performing both inverse and forward modeling with quantified uncertainty, even under extremely sparse observational conditions. These results underscore SURGIN’s efficacy and adaptability in zero-shot reconstruction of geological fields and the spatiotemporal dynamics of state variables from sparse well measurements, highlighting its ability to handle diverse monitoring setups without the need for configuration-specific retraining.

\subsubsection{Super-resolution for low-fidelity observations}

\begin{figure}[htb!]
\centering
\includegraphics[width=\textwidth]{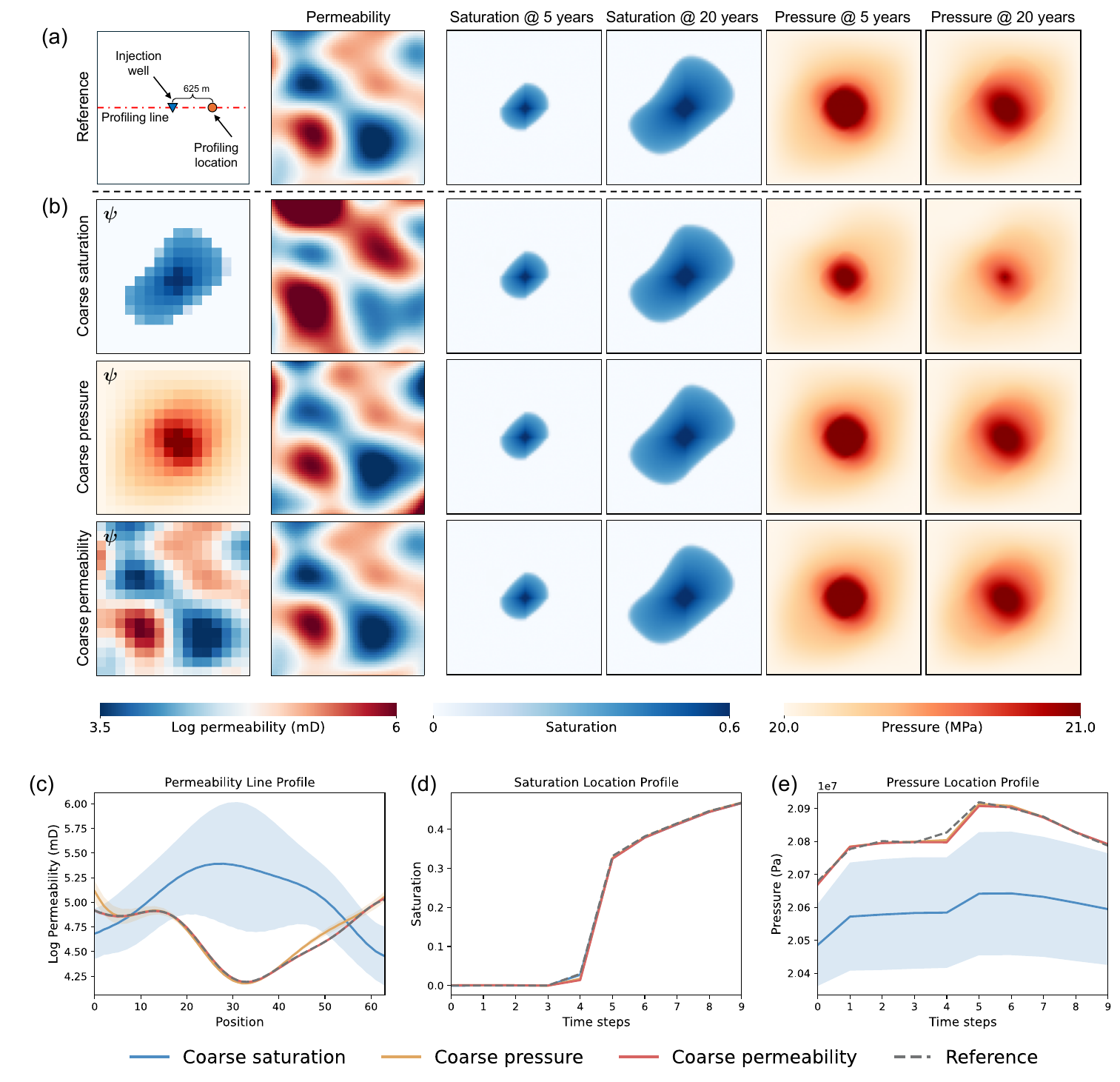}
\caption{(a) Reference permeability fields and state variables ($\mathrm{CO_2}$ saturation and pressure) at 5 and 20 years. (b) Generated permeability fields and predicted state variables conditioned on coarse observations of saturation, pressure or permeability. (c) Permeability profiles along the profiling line. (d) Temporal profiles of saturation at the profiling location. (e) Temporal profiles of pressure at the same location.}\label{fig_case1_sr1}
\end{figure}

\begin{figure}[htb!]
\centering
\includegraphics[width=0.8\textwidth]{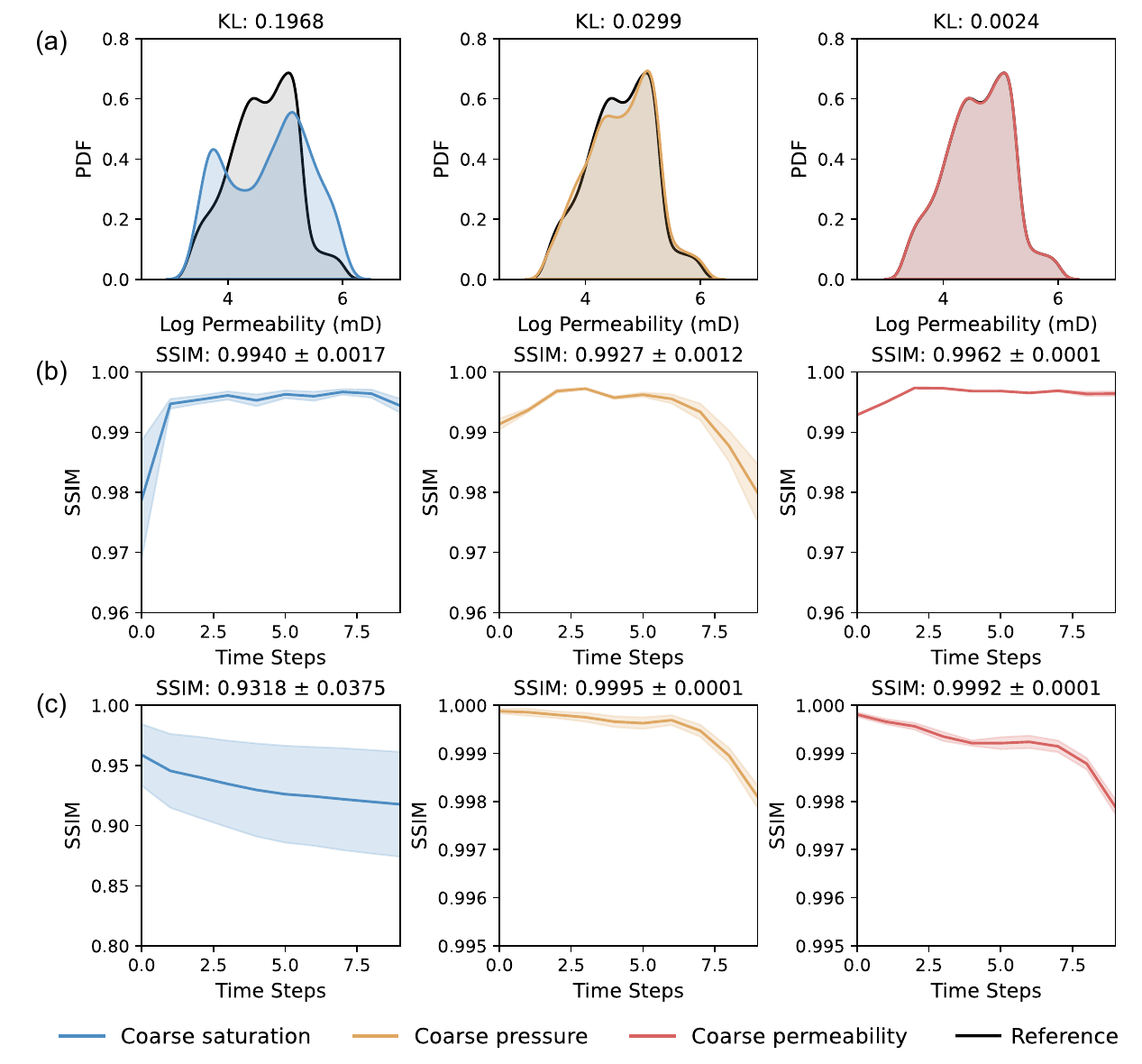}
\caption{Quantitative metrics evaluated over all generated ensemble samples conditioned on the coarse observations. (a) PDF of the generated permeability fields compared with the reference, with the corresponding KL divergence. (b) SSIM of saturation between generated and reference trajectories, with shaded areas denoting the standard deviation. (c) SSIM of pressure between generated and reference trajectories.}\label{fig_case1_sr2}
\end{figure}

In field applications of subsurface multiphase flow engineering, the interpreted geological fields and observed reservoir responses can be under-resolved owing to the resolution limits of non-intrusive measurements such as seismic monitoring or to incomplete field records~\cite{reynolds_introduction_nodate,eid_seismic_2015,ajayi_review_2019}. We therefore investigate SURGIN’s inversion performance under low-fidelity observational scenarios, with emphasis on low-resolution data $\bm\psi$ arising from either the solution space or the parameter space. As depicted in Figure~\ref{fig_case1_sr1}a and b, the permeability fields inferred solely from low-resolution $\mathrm{CO_2}$ time-lapse seismic monitoring deviate substantially from the reference field, although the inferred saturation dynamics remains consistent with the reference owing to the availability of direct observations in that functional space (Figure~\ref{fig_case1_sr1}d). This is because the complete geological field cannot be reliably reconstructed from localized observations of confined $\mathrm{CO_2}$ saturation maps. In contrast, when observations of coarse pressure or permeability fields are available, which exhibit global spatial patterns, the inversion achieves high fidelity, yielding consistency with the ground truth in both geological parameters and the corresponding state variables (Figure~\ref{fig_case1_sr1}c, d and e). This can be further corroborated through the quantitative analysis in Figure~\ref{fig_case1_sr2}, which shows that direct observations of pressure and permeability lead to superior inversion and prediction performance, with KL divergence of the geological fields below 0.03 and SSIM of the state variables exceeding 0.98. In general, SURGIN demonstrates the capability to address super-resolution tasks in geoscientific modeling, reconstructing fine-scale geological parameter fields and resolving high-resolution spatiotemporal dynamics of state variables. This highlights its versatility in bridging scales and improving the fidelity of subsurface characterization.

\subsection{Case 2: vertical model with post-injection}
\subsubsection{Numerical model setup}
Case 2 investigates a distinct subsurface multiphase flow scenario in which supercritical $\mathrm{CO_2}$ is injected through a vertical well followed by a post-injection period, highlighting the interplay between buoyancy and lateral driving forces (e.g., viscous and capillary effects), further compounded by structural heterogeneity of the reservoir. The rectangular reservoir domain extends $5~\mathrm{km}\times100~\mathrm{m}\times100~\mathrm{m}$ and is discretized into $64\times1\times64$ uniform grids, serving as a quasi 2-dimensional model that represents a vertical slice of the reservoir. The vertical well, located at the horizontal midpoint of the domain, is perforated across all reservoir layers and operates at a constant injection rate of 4~kg/s. The left and right boundaries are modeled as open-flow boundaries to mimic connection to an infinite hydrosystem, whereas the top and bottom boundaries are treated as closed, representing impermeable bedrock and caprock layers. The injection period lasts for 4 years, followed by a 6-year post-injection phase during which the gaseous $\mathrm{CO_2}$ plume gradually dissolves into the formation water, governed by complex interactions between gravitational segregation and convection-dissolution processes. The outputs are collected annually, resulting in a total 10 time steps. Heterogeneous permeability fields, drawn from log-normal Gaussian distributions, are imposed on the reservoir flow properties, serving as a stochastic representation of subsurface heterogeneity and enabling a wide range of geological realizations. The horizontal and vertical correlation lengths are set to 1562.5~m and 6.25~m, respectively, to emphasize the stronger heterogeneity in the vertical direction, which is a characteristic feature of sedimentary rock reservoirs. The other numerical parameters, including relative permeability and capillary relations, are identical to those in case 1.

Case 2 explores a series of more challenging inversion tasks with the aim of demonstrating SURGIN’s capability to assimilate diverse modalities of observational information $\bm\psi$. In one scenario, the goal is to infer permeability fields using data from only three monitoring wells, with records available solely during the injection period. This constitutes a history matching task, where historical data are assimilated during model calibration to reconstruct the geological model and subsequently forecast future reservoir states. We also account for noise perturbations in this scenario, reflecting the conditions typically encountered in engineering applications. In another scenario, we aim to reconstruct the full fields of either geological parameters $\mathbf{m}$ or state variables $\mathbf{u}$ from spatially masked records, addressing challenging inpainting tasks that are common when field measurements or numerical simulations contain missing data.

\subsubsection{History matching with temporally incomplete records}

\begin{figure}[htb!]
\centering
\includegraphics[width=\textwidth]{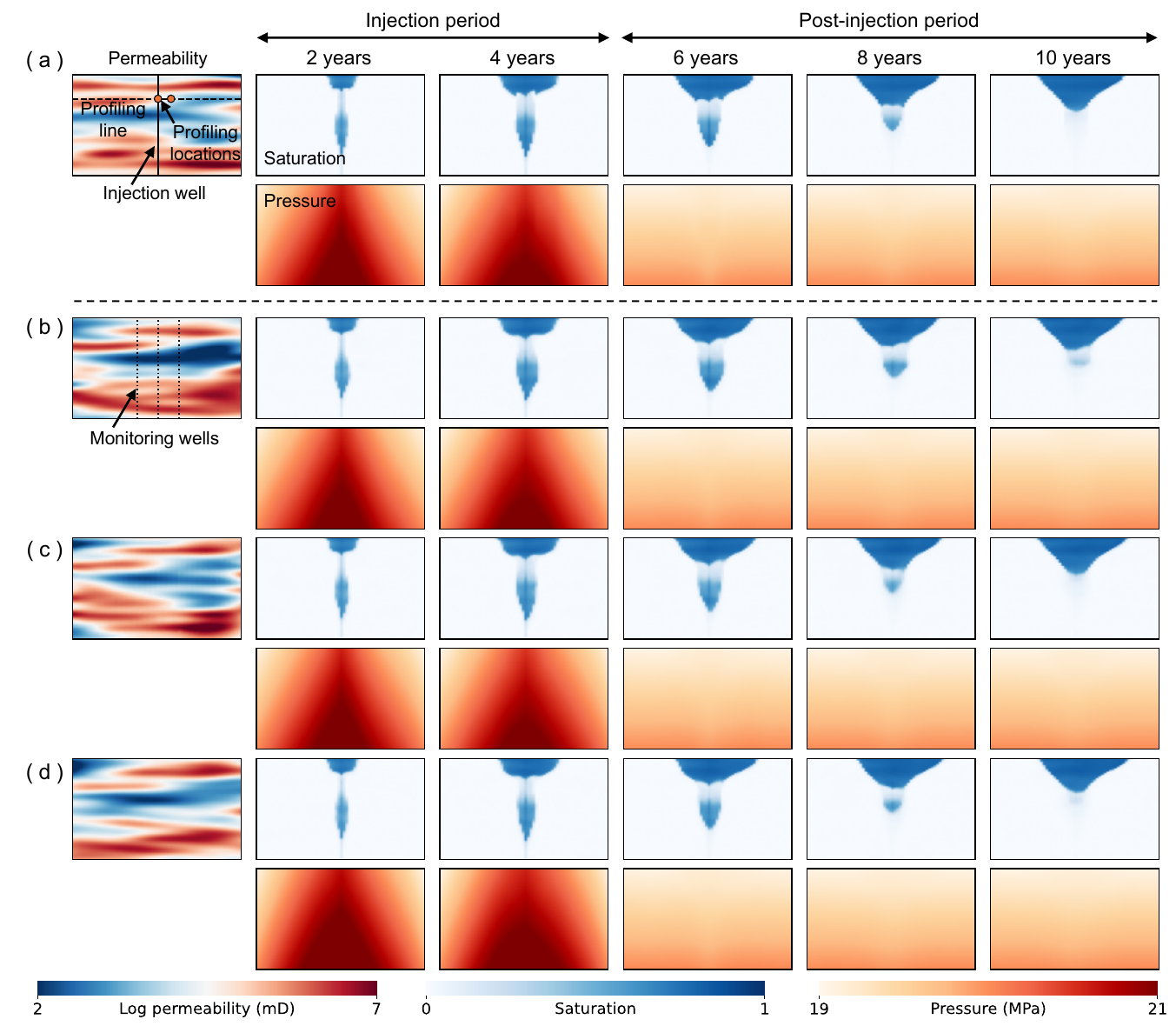}
\caption{(a) Reference permeability fields and state variables ($\mathrm{CO_2}$ saturation and pressure) at 2, 4, 6, 8, and 10 years. The first 4 years correspond to the injection period, followed by 6 years of post-injection. (b–d) Generated permeability fields and predicted state variables conditioned on sparse well measurements under three scenarios: (b) state variables only (solution), (c) both state variables and permeability values (solution + parameter), and (d) noisy state variables (noisy solution).}\label{fig_case2_sp1}
\end{figure}

\begin{figure}[htb!]
\centering
\includegraphics[width=0.6\textwidth]{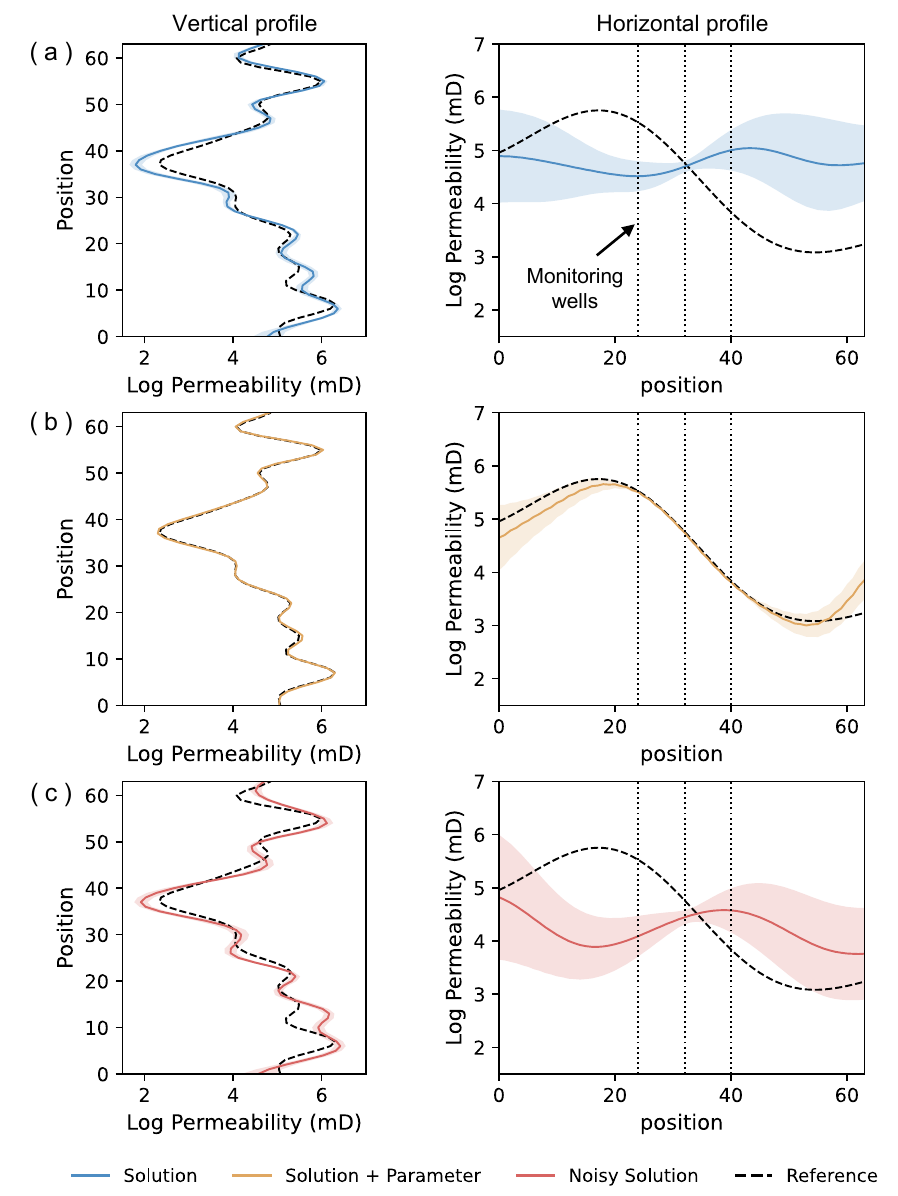}
\caption{Vertical (injection well/central monitoring well) and horizontal (profiling line) profiles of the permeability fields. Dotted lines indicate monitoring well locations, and shaded areas denote uncertainty.}\label{fig_case2_sp2}
\end{figure}

\begin{figure}[htb!]
\centering
\includegraphics[width=\textwidth]{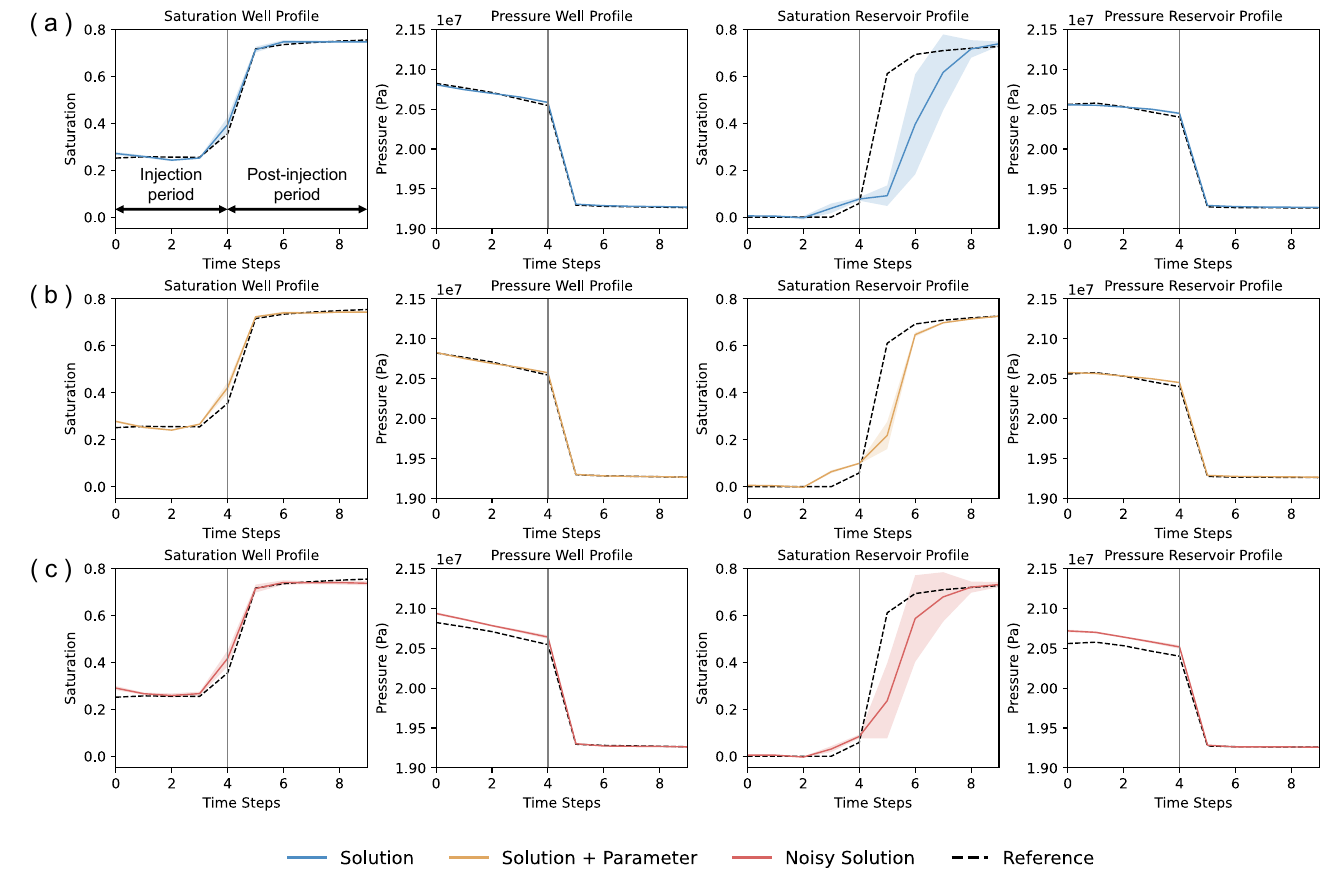}
\caption{Temporal profiles of state variables at well and reservoir profiling locations. Shaded areas denote uncertainty. Vertical gray lines separate the injection and post-injection period.}\label{fig_case2_sp3}
\end{figure}

\begin{figure}[htb!]
\centering
\includegraphics[width=0.8\textwidth]{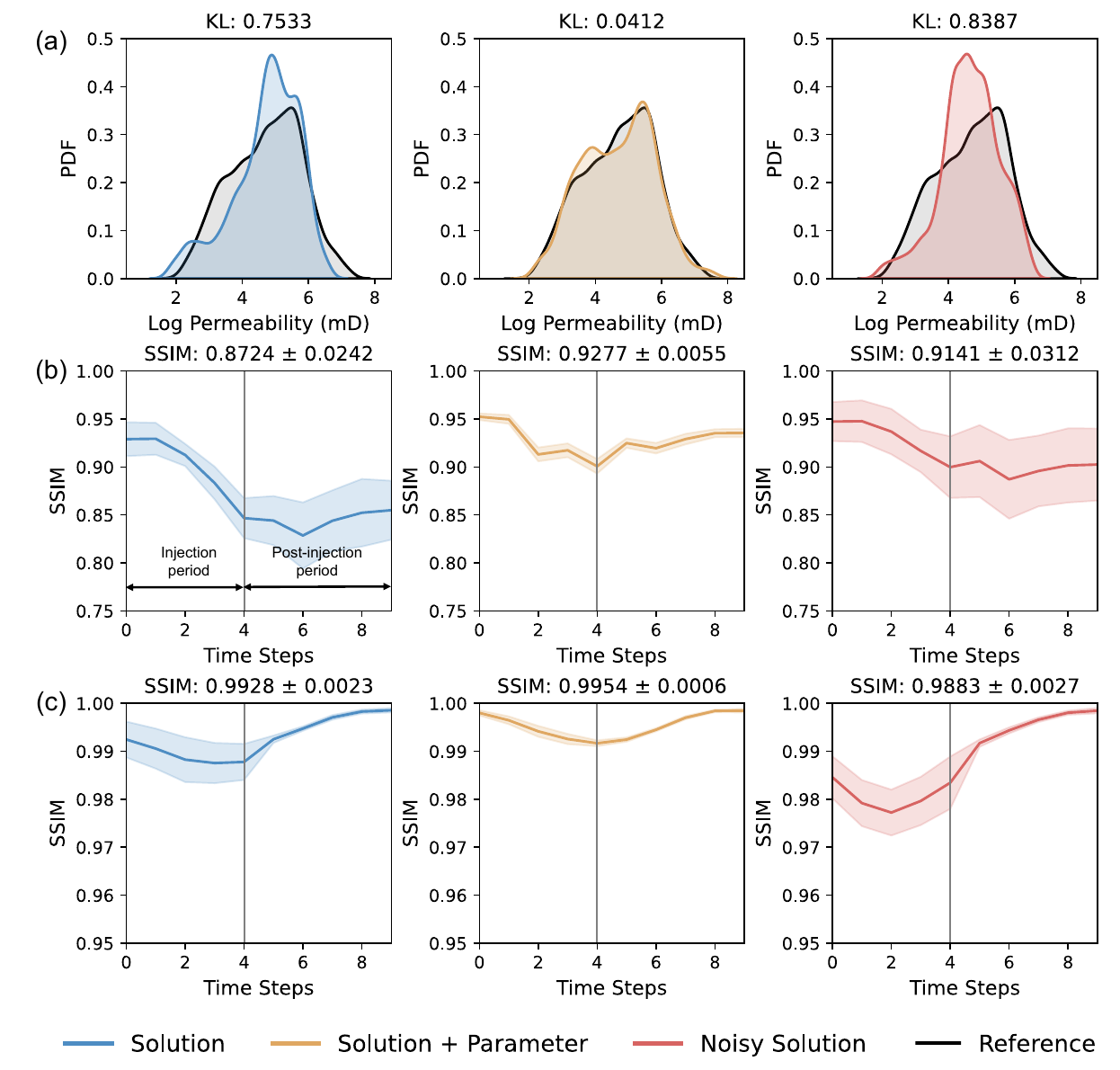}
\caption{Quantitative metrics evaluated over all generated ensemble samples conditioned on the sparse vertical well measurements. (a) PDF of the generated permeability fields compared with the reference, with the corresponding KL divergence. (b) SSIM of saturation between generated and reference trajectories, with shaded areas denoting the standard deviation. (c) SSIM of pressure between generated and reference trajectories.}\label{fig_case2_sp4}
\end{figure}

We consider sparse well measurements obtained from 3 monitoring wells: the injection well itself and two symmetric vertical wells located 625~m away from it. Unlike the sparse well measurements in case 1, where records are available at all time steps, here observations are limited to the injection period. Thus, the task involves history matching during the injection phase, while the post-injection phase is treated purely as a forecast, framing the problem as inverse modeling with temporally incomplete records. From the reference geological parameters and the corresponding $\mathrm{CO_2}$ saturation and pressure dynamics in Figure~\ref{fig_case2_sp1}a, it can be observed that the heterogeneity is stronger than in case 1. The saturation initially expands and is shaped by the heterogeneous geological patterns, then migrates upward beneath the caprock and becomes laterally flattened during the post-injection phase. Meanwhile, the pressure exhibits overpressurization during injection and gradually returns to its in-situ value once injection ceases. These spatial heterogeneities and spatiotemporal dynamics, though complex, can yet be faithfully inferred from extremely sparse measurements using our proposed SURGIN framework. We scrutinize three types of observational conditions: (1) solution measurements (saturation and pressure), (2) combined solution and parameter measurements (including permeability values), and (3) noisy solution measurements. The corresponding inversion results are presented in Figure~\ref{fig_case2_sp1}b–d, respectively. Overall, SURGIN can effectively delineate zones of elevated and reduced permeability within the reservoir, although certain fine-scale details remain unresolved. Notably, when direct permeability values are incorporated, the inferred geological fields exhibit the closest alignment with the reference. The spatial profiles of the geological fields further substantiate these findings, as the uncertainty is drastically reduced when conditioned on both the solution and parameter spaces (Figure~\ref{fig_case2_sp2}). The predicted spatiotemporal dynamics of the state variables align closely with the reference during the history matching phase, whereas discrepancies and uncertainties emerge in the post-injection forecast phase, especially for scenarios involving solution-only and noisy observations (Figure~\ref{fig_case2_sp3}). The above trends are further corroborated by the quantitative metrics in Figure~\ref{fig_case2_sp4}. Interestingly, during the post-injection period, while pressure dynamics are comparatively easier to predict — since diffusive overpressurization relaxes to a quasi-static equilibrium that is inherently simple to capture — the saturation dynamics are considerably more challenging to reproduce, as they are controlled by the intricate interplay of gravitational segregation, buoyancy-driven migration, and convection-dissolution processes. (Figure~\ref{fig_case2_sp4}b and c).

\subsubsection{Inpainting of spatially corrupted data}

\begin{figure}[htb!]
\centering
\includegraphics[width=0.95\textwidth]{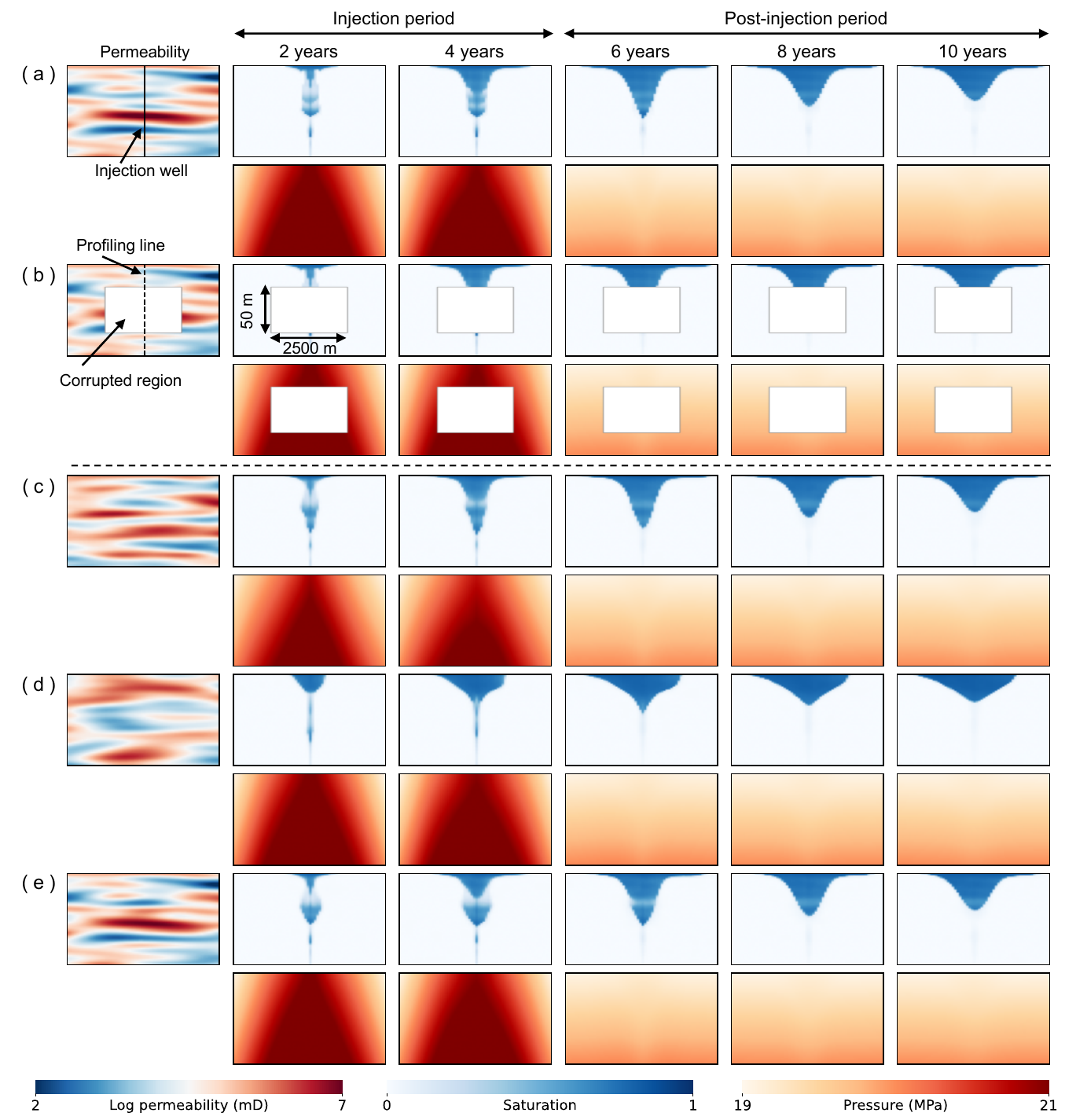}
\caption{(a) Reference permeability fields and state variables ($\mathrm{CO_2}$ saturation and pressure) at 2, 4, 6, 8, and 10 years. The first 4 years correspond to the injection period, followed by 6 years of post-injection. (b) The corrupted region is a 2500~m~$\times$~50~m subregion at the center of the fields, mathematically formulated as a spatial or spatiotemporal masking operation. (c–e) Generated permeability fields and predicted state variables conditioned on incomplete observations under three scenarios: (c) corrupted saturation, (d) corrupted pressure, and (d) corrupted permeability.}\label{fig_case2_inp1}
\end{figure}

\begin{figure}[htb!]
\centering
\includegraphics[width=0.75\textwidth]{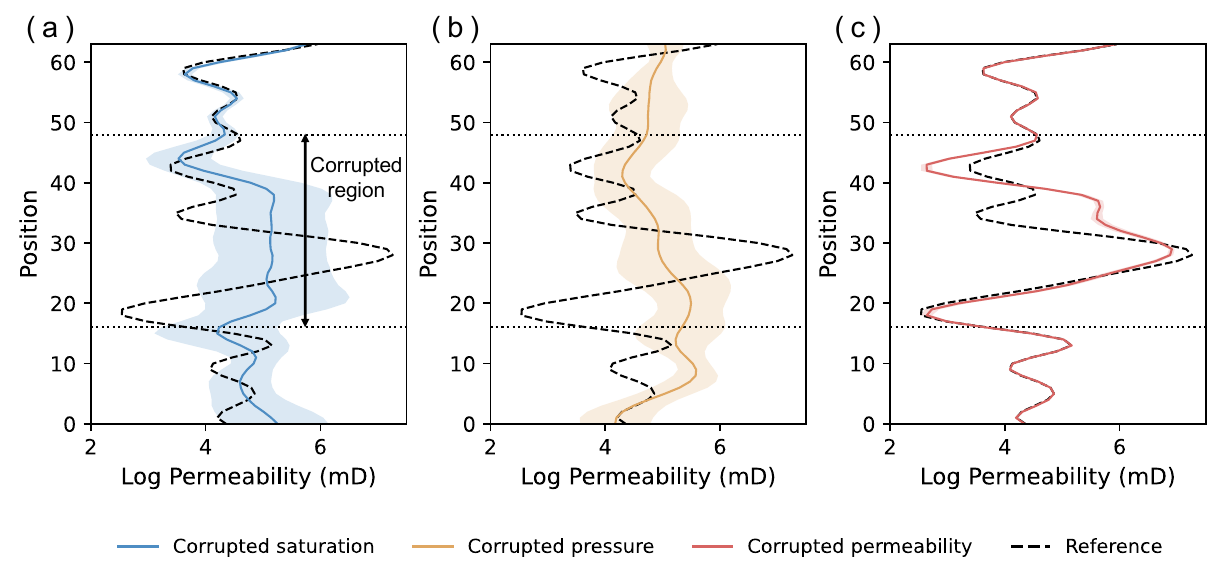}
\caption{Vertical profiles of the permeability fields along the profiling line. Shaded areas represent uncertainty, and regions between the dotted lines indicate the corrupted domain.}\label{fig_case2_inp2}
\end{figure}

\begin{figure}[htb!]
\centering
\includegraphics[width=0.75\textwidth]{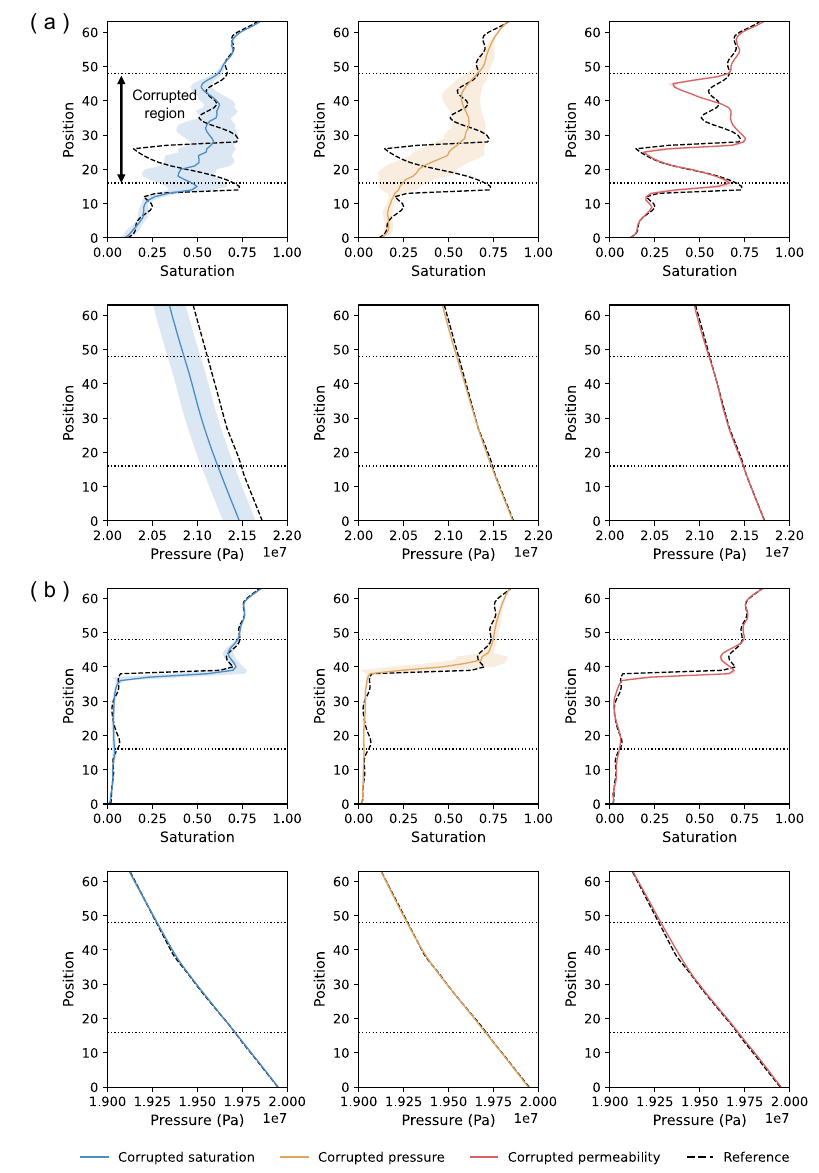}
\caption{Spatial profiles of state variables at (a) 4 years and (b) 10 years. Shaded areas represent uncertainty, and regions between the dotted lines indicate the corrupted domain.}\label{fig_case2_inp3}
\end{figure}

\begin{figure}[htb!]
\centering
\includegraphics[width=0.8\textwidth]{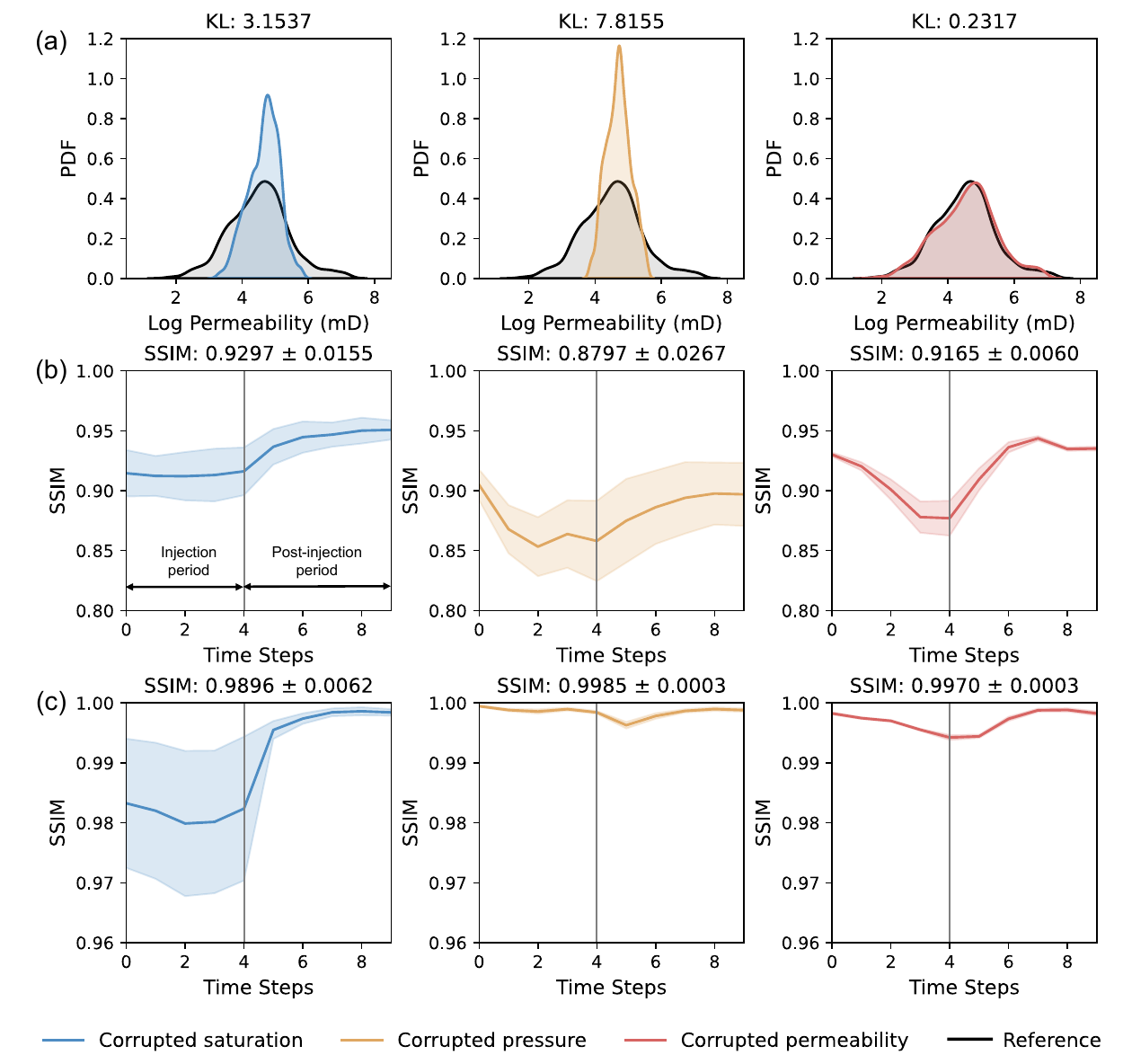}
\caption{Quantitative metrics evaluated over all generated ensemble samples conditioned on the spatially corrupted data. (a) PDF of the generated permeability fields compared with the reference, with the corresponding KL divergence. (b) SSIM of saturation between generated and reference trajectories, with shaded areas denoting the standard deviation. (c) SSIM of pressure between generated and reference trajectories.}\label{fig_case2_inp4}
\end{figure}

Field monitoring data may be partially corrupted in specific regions as a result of equipment malfunction, limited geological surveying, or missing historical subsurface records~\cite{reynolds_introduction_nodate,mahjour_risks_2023}. We therefore explore the capacity of SURGIN to address data restoration problems, analogous to image inpainting~\cite{yu_generative_2018} for static geological parameters and video inpainting~\cite{xu_deep_2019} for spatiotemporal state variable dynamics, within complex subsurface multiphase systems. Figure~\ref{fig_case2_inp1} presents another representative test case of this challenging inverse problem, where the conditioning data $\bm\psi$ consist of the fields excluding the central masked-out subregion (2500~m~$\times$~50~m). SURGIN is employed to recover the lost information and reconstruct the full fields via conditional generation. Figure~\ref{fig_case2_inp1}c–d compares inversion results when the input conditions are corrupted saturation, pressure, or permeability fields, respectively. These visualizations reveal that when direct observations are available in a functional space, the inferred full fields within that space become accurate, since the inversion no longer relies solely on indirect correlations but is directly constrained by the observed information. The permeability profiles along the injection well trajectory demonstrate that SURGIN can faithfully reconstruct the full permeability field from its corrupted counterparts, with only minor discrepancies within the masked region (Figure~\ref{fig_case2_inp2}). Moreover, Figure~\ref{fig_case2_inp3} shows that the associated uncertainty increases from the periphery toward the center of the masked area. This pattern arises because grid points near the edges exhibit stronger spatial covariance with adjacent known regions, thereby reducing uncertainty relative to the central portion of the masked domain. Further evidence of SURGIN's proficiency is presented in Figure~\ref{fig_case2_inp4}, which shows that the model can satisfactorily infer complete fields from their masked versions.

\section{Discussion}\label{sec5}

In contrast to conventional inversion methods, which either rely on pre-curated datasets tied to specific conditions~\cite{wang_generative_2025,zhan_toward_2025} or require rerunning the entire inversion algorithm when new conditions arise~\cite{han_surrogate_2024,jiang_history_2024,seabra_ai_2024}, the proposed SURGIN framework is pretrained only once and can then conditionally generate posterior samples under arbitrary unseen observations without further retraining at inference. In Table~\ref{table_comp_time}, we present the computational time of conditional generation for the numerical experiments. Across all observation scenarios, SURGIN requires less than one minute to generate an ensemble of posterior realizations, enabled by the rapid inference speed of the surrogate model and the parallel computing capability of GPUs. Compared with numerical forward simulation, which can take more than 10 minutes per realization and becomes even more costly for traditional inverse algorithms that require repeated forward evaluations, our direct inversion method achieves a substantial speedup. This feature holds substantial practical value, providing an efficient pathway for real-time data assimilation in subsurface multiphase flow systems.

\begin{table}[htb!]
\caption{Computational time of conditional generation and numerical forward simulation in the case studies.}\label{table_comp_time}
\centering
\begin{tabularx}{\textwidth}{@{} c X c c @{}}
\toprule
Case & Conditioning scenario & Generation (sec/batch) & Simulation (min/sample)\\
\midrule
\multirow{3}{*}{1}
& Sparse well              & 45 & \multirow{3}{*}{$\sim$15}\\
\cmidrule(lr){2-3}
& Coarse parameters        & 14 & \\
\cmidrule(lr){2-3}
& Coarse solutions         & 28 & \\

\midrule

\multirow{3}{*}{2}
& Sparse well              & 45 & \multirow{3}{*}{$\sim$10}\\
\cmidrule(lr){2-3}
& Corrupted parameters     & 14 & \\
\cmidrule(lr){2-3}
& Corrupted solutions      & 28 & \\

\bottomrule
\end{tabularx}
\end{table}

\begin{figure}[htb!]
\centering
\includegraphics[width=\textwidth]{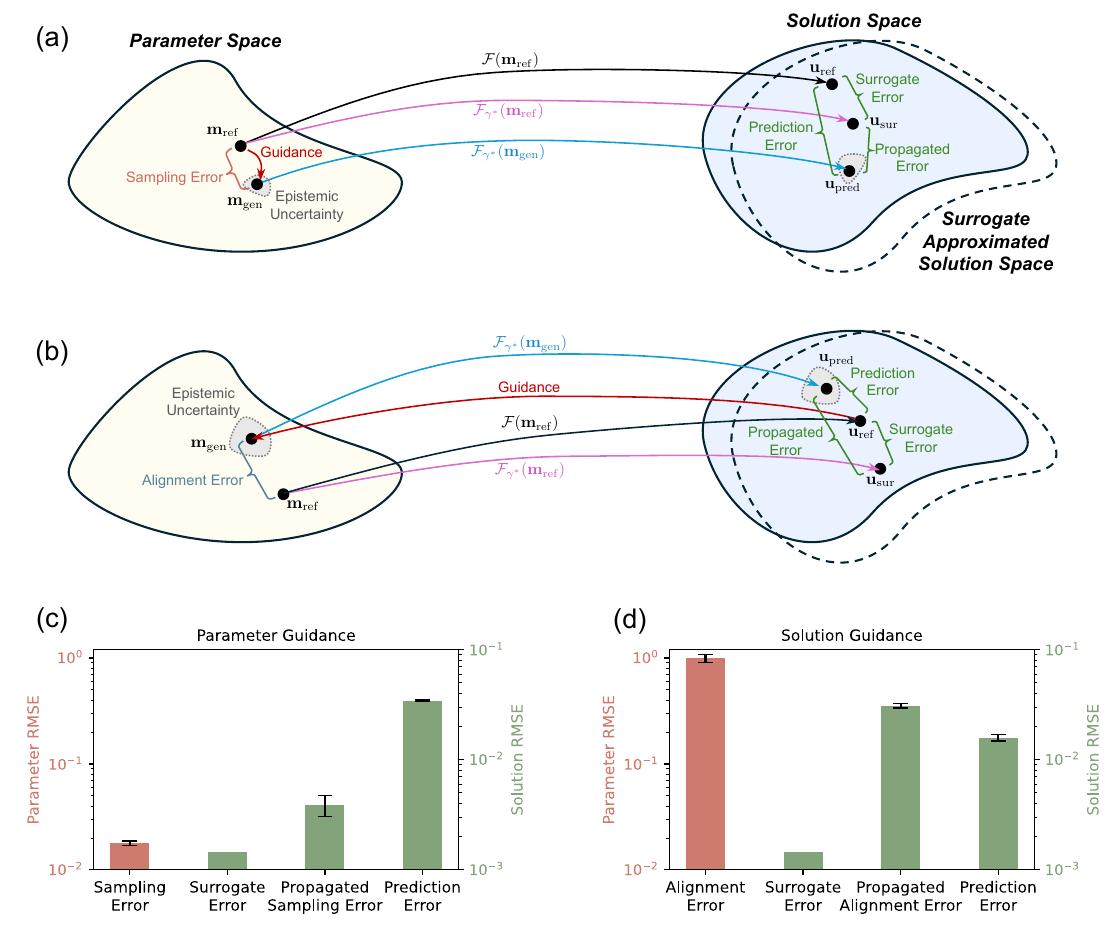}
\caption{Error and uncertainty propagation in the SURGIN framework, where guidance corresponds to full observations of the (a) parameter space and (b) solution space. (c–d) Errors quantified by the root mean square error (RMSE): sampling error denotes the discrepancy between generated $\mathbf{m}_{\mathrm{gen}}$ and reference $\mathbf{m}_{\mathrm{ref}}$ permeability fields under full permeability observation; alignment error denotes the discrepancy between $\mathbf{m}_{\mathrm{gen}}$ and $\mathbf{m}_{\mathrm{ref}}$ under full solution observation; surrogate error denotes the discrepancy between $\mathbf{u}_{\mathrm{sur}}=\mathcal{F}_{\gamma^*}(\mathbf{m}_{\mathrm{ref}})$ and the high-fidelity reference solution $\mathbf{u}_{\mathrm{ref}}=\mathcal{F}(\mathbf{m}_{\mathrm{ref}})$, capturing the bias of the surrogate approximation; propagated sampling/alignment error refers to the discrepancy between $\mathbf{u}_{\mathrm{pred}}=\mathcal{F}_{\gamma^*}(\mathbf{m}_{\mathrm{gen}})$ and $\mathbf{u}_{\mathrm{sur}}=\mathcal{F}_{\gamma^*}(\mathbf{m}_{\mathrm{ref}})$, quantifying how parameter discrepancies (sampling or alignment) are transmitted into the solution space; and prediction error is defined as the discrepancy between the surrogate-predicted solution from the generated parameters, $\mathbf{u}_{\mathrm{pred}}$, and the high-fidelity reference solution, $\mathbf{u}_{\mathrm{ref}}$. All metrics are evaluated on the final numerical sample.}\label{fig_discuss}
\end{figure}

We further discuss the sources of uncertainty and error within the SURGIN framework. In the preceding numerical experiments, the pretrained unconditional SGM serves as a prior, encapsulating epistemic uncertainty associated with the calibration of a specific geological realization. This form of uncertainty can be reduced as the amount of observational information increases. In addition, aleatoric uncertainty, originating from measurement noise, can propagate into the posterior outputs through the mapping $\mathcal{M}$, and is irreducible. However, the epistemic uncertainty associated with neural network parameters is not explicitly accounted in the current work. We now consider the case where full observations of either the geological parameters or the state variables are available, where the epistemic uncertainty arising from incomplete observations vanishes. Figure~\ref{fig_discuss} illustrates the origins of error and uncertainty in SURGIN, using the final numerical sample as a representative case. Sampling error arises in parameter-guided generation because Bayesian posterior sampling is inherently stochastic. Even with a well-specified prior, the reverse SDE may follow different trajectories, and the sampling procedure itself may become trapped in local minima. This error, albeit small, embodies the epistemic uncertainty rooted in the stochastic nature of posterior exploration, as indicated by the standard deviation of sampling error in Figure~\ref{fig_discuss}c. Once the generated parameters are used to predict solutions through the surrogate model, prediction error inevitably emerges, with epistemic uncertainties from sampling further carried into the predicted solutions, as reflected by the standard deviation of prediction error in Figure~\ref{fig_discuss}c. In addition, we define a propagated error, which is measured by the discrepancy between $\mathbf{u}_{\mathrm{pred}}=\mathcal{F}_{\gamma^*}(\mathbf{m}_{\mathrm{gen}})$ and $\mathbf{u}_{\mathrm{sur}}=\mathcal{F}_{\gamma^*}(\mathbf{m}_{\mathrm{ref}})$. The propagated error isolates the portion of solution discrepancy that originates solely from the stochasticity of parameter generation. In other words, it quantifies how errors in the sampled geological parameters are transmitted into the solution space when both $\mathbf{m}_{\mathrm{gen}}$ and $\mathbf{m}_{\mathrm{ref}}$ evaluated with the same surrogate model, thereby disentangling surrogate-induced bias from sampling-induced error. In the case of solution-guided generation ( Figure~\ref{fig_discuss}b), the error sources manifest differently. Because the generative process is constrained directly by the reference solution, the predicted solution $\mathbf{u}_{\mathrm{pred}}$ generally aligns well with $\mathbf{u}_{\mathrm{ref}}$, resulting in a relatively small prediction error. However, this guidance does not enforce consistency in the parameter space, and the accumulated approximation errors of the surrogate forward operator during guided sampling can amplify discrepancies between the generated and reference parameters. We term this discrepancy the alignment error, which reflects the misalignment between parameter and solution spaces. As a direct consequence, the propagated error, which depends on parameter differences, can exceed the prediction error, since $\mathbf{u}_{\mathrm{sur}}$ remains tied to the reference parameters (Figure~\ref{fig_discuss}d). This inversion of relative magnitudes reflects the fact that solution-guided generation prioritizes fidelity in the solution space, even at the expense of parameter consistency.

\section{Conclusions}\label{sec6}
We have presented SURGIN, an innovative probabilistic generative framework for conditional generation of subsurface geological parameters together with the corresponding spatiotemporal reservoir state variables. At its core, SURGIN uniquely couples a U-FNO surrogate with an SGM, enabling efficient and observation-consistent generation. The SGM parameterizes the prior distribution of geological realizations through self-supervised pretraining, while posterior samples are inferred via surrogate-guided Bayesian conditional sampling without model retraining. SURGIN has demonstrated its proficiency in inverse modeling across diverse monitoring configurations in complex subsurface flow scenarios, finding applicability in full-field reconstruction from sparse well measurements, super-resolution of coarsely observed data, and restoration of corrupted inputs. The results reveal that SURGIN achieves sound accuracy, generalizability, and robustness in inferring geological fields and predicting spatiotemporal reservoir dynamics with quantified uncertainty, even under severe data sparsity and noise. In summary, our effort establishes a new class of direct inversion methods for subsurface multiphase flow data assimilation, marking a significant advancement in generative learning within parametric functional spaces.

While the SURGIN framework shows promise for zero-shot characterization of complex subsurface multiphase flow systems, it remains computationally demanding in high-dimensional geological settings, particularly when 3-dimensional or unstructured geomodels are involved. This challenging, however, can be mitigated through appropriate dimensionality reduction techniques and adapted patching strategies. Moreover, as a purely data-driven method, SURGIN requires rich training data. Nevertheless, this limitation is expected to be alleviated by incorporating explicit physical constraints during surrogate pretraining and posterior sampling, or by adopting differentiable forward solvers for likelihood evaluation, which may also help resolve the functional space misalignment issue discussed above. Looking forward, we aim to systematically address these potential limitations, paving the way toward a new generation of uncertainty-aware, physics-informed generative inversion methods for complex subsurface systems.

\newpage
\appendix
\section{Training and inference protocols of SURGIN}\label{hyperparam}
To facilitate training, the continuous SDEs are discretized into 1,000 steps for forward diffusion and reverse denoising. During SGM pretraining, noise levels are sampled from a cosine noise schedule and added to the data, with the DiT score network trained to predict the injected noise. The learning rate is set to $5\times10^{-5}$, the batch size to 64, and training proceeds for 350,000 iterations until loss convergence. An exponential moving average with a decay rate of 0.9999 is applied for model saving.

For the U-FNO surrogate, the learning rate is 0.001 with a step-LR scheduler (step size = 2, multiplicative factor of decay = 0.9), a batch size of 50, and 150 training epochs. Training uses an $L_p$-relative loss with an additional spatial gradient penalty (coefficient = 0.5). All pretraining experiments are conducted on a single NVIDIA RTX 4090 GPU.

For the online inference, the step size of guidance ${1}/{\sigma_c}$ is set to 1. Posterior sampling is performed with an Adam-based update rule. The scale parameter is set to 0.01, with the first-moment decay rate set to 0.9, the second-moment decay rate set to 0.999, and the numerical stability constant set to $1\times10^{-8}$.

\section{U-FNO architecture}\label{ufno}

\begin{table}[H]
\caption{U-FNO model architecture. The Padding layer is used to accommodate the non-periodic boundaries. The output tensor comprises four channels: the first two encode the 2D spatial grids, the third encodes time, and the fourth encodes feature values. The batch channel is omitted for clarity. The input features include the permeability field and the physical timestep.}
\centering
\begin{tabularx}{\textwidth}{@{}p{4cm} X l@{}}
\toprule
Name & Layer  & Output shape\\
\midrule
Input    & -   & (64, 64, 10, 2)\\
Padding    & Padding   & (72, 72, 18, 2)\\
Lifting    & Linear   & (72, 72, 18, 36)\\
Fourier 1   & Fourier3d/Conv1d/Add/ReLu & (72, 72, 18, 36)\\
Fourier 2   & Fourier3d/Conv1d/Add/ReLu & (72, 72, 18, 36)\\
Fourier 3   & Fourier3d/Conv1d/Add/ReLu & (72, 72, 18, 36)\\
U-Fourier 1   & Fourier3d/Conv1d/UNet3d/Add/ReLu & (72, 72, 18, 36)\\
U-Fourier 2   & Fourier3d/Conv1d/UNet3d/Add/ReLu & (72, 72, 18, 36)\\
U-Fourier 3   & Fourier3d/Conv1d/UNet3d/Add/ReLu & (72, 72, 18, 36)\\
Projection 1    & Linear    & (72, 72, 18, 128)\\
Projection 2    & Linear    & (72, 72, 18, 1)\\
De-padding  & -     & (64, 64, 10, 1)\\
\bottomrule
\end{tabularx}
\end{table}

\section{Diffusion Transformer architecture}\label{dit}

\begin{table}[H]
\caption{DiT score network architecture. The input is a single-channel noisy permeability field with a spatial resolution of 64~$\times$~64, which is first embedded into a sequence of length 1024 with a hidden size of 256 using a patch size of 2. The embedded sequence is then processed by 12 DiT blocks, followed by adaptive normalization and a linear projection, before being reshaped back to its original resolution to produce the added noise.}
\centering
\begin{tabularx}{\textwidth}{@{}p{4cm} X l@{}}
\toprule
Name & Layer  & Output shape\\
\midrule
Input    & -   & (1, 64, 64)\\
Embedding    & Patchify   & (1024, 256)\\
Processing 1    & DiT Block   & (1024, 256)\\
Processing 2    & DiT Block   & (1024, 256)\\
Processing 3    & DiT Block   & (1024, 256)\\
Processing 4    & DiT Block   & (1024, 256)\\
Processing 5    & DiT Block   & (1024, 256)\\
Processing 6    & DiT Block   & (1024, 256)\\
Processing 7    & DiT Block   & (1024, 256)\\
Processing 8    & DiT Block   & (1024, 256)\\
Processing 9    & DiT Block   & (1024, 256)\\
Processing 10    & DiT Block   & (1024, 256)\\
Processing 11    & DiT Block   & (1024, 256)\\
Processing 12    & DiT Block   & (1024, 256)\\
Normalization    & Layer Norm   & (1024, 256)\\
Projection    & Linear   & (1024, 4)\\
Reshape  & -     & (1, 64, 64)\\
\bottomrule
\end{tabularx}
\end{table}

\section{Proof of the equivalence between $\mathcal{L}^{\mathrm{SM}}$ and $\mathcal{L}^{\mathrm{DSM}}$}\label{sm_loss}
The original score matching loss function is given by,
\begin{align}
    \mathcal{L}^{\mathrm{SM}}(\theta)&=\mathbb{E}_{p(\tau)}\bigg[\lambda(\tau)\mathbb{E}_{p(\mathbf{m}(\tau))}\big[\|\mathbf{s}_\theta(\mathbf{m}(\tau),\tau)-\nabla_{\mathbf{m}(\tau)}\log p(\mathbf{m}(\tau)) \|^2\big]\bigg] \nonumber\\
    &=\mathbb{E}_{p(\tau)}\bigg[\lambda(\tau)\mathbb{E}_{p(\mathbf{m}(\tau))}\big[\|\mathbf{s}_\theta(\mathbf{m}(\tau),\tau)\|^2+\|\nabla_{\mathbf{m}(\tau)}\log p(\mathbf{m}(\tau)) \|^2-\nonumber\\ &\ \ \ \ \ \mathbf{s}_\theta(\mathbf{m}(\tau),\tau)^T\nabla_{\mathbf{m}(\tau)}\log p(\mathbf{m}(\tau))\big]\bigg],
\end{align}
where the second term $\|\nabla_{\mathbf{m}(\tau)}\log p(\mathbf{m}(\tau)) \|^2\coloneqq C_1$ is a constant independent of $\theta$, and the last term can be developed by using the definition of expectation and chain rule,
\begin{align}\label{b2}
    \mathbb{E}_{p(\mathbf{m}(\tau))}[\mathbf{s}_\theta(\mathbf{m}(\tau),\tau)^T&\nabla_{\mathbf{m}(\tau)}\log p(\mathbf{m}(\tau))]\nonumber\\
    &=\int\bigg[\mathbf{s}_\theta(\mathbf{m}(\tau),\tau)^T\nabla_{\mathbf{m}(\tau)}\log p(\mathbf{m}(\tau))\bigg]p(\mathbf{m}(\tau))d\mathbf{m}(\tau)\nonumber\\
    &=\int\bigg[\mathbf{s}_\theta(\mathbf{m}(\tau),\tau)^T\frac{\nabla_{\mathbf{m}(\tau)} p(\mathbf{m}(\tau))}{p(\mathbf{m}(\tau))}\bigg]p(\mathbf{m}(\tau))d\mathbf{m}(\tau)\nonumber\\
    &=\int\mathbf{s}_\theta(\mathbf{m}(\tau),\tau)^T\nabla_{\mathbf{m}(\tau)} p(\mathbf{m}(\tau))d\mathbf{m}(\tau).
\end{align}

Recalling the marginal distribution,
\begin{equation}
    p(\mathbf{m}(\tau))=\int p(\mathbf{m}(\tau)|\mathbf{m}(0))p(\mathbf{m}(0))d\mathbf{m}(0),
\end{equation}
we can rewrite Eq.~\ref{b2} as,
\begin{align}\label{b4}
    \int\mathbf{s}_\theta&(\mathbf{m}(\tau),\tau)^T\nabla_{\mathbf{m}(\tau)} p(\mathbf{m}(\tau))d\mathbf{m}(\tau)\nonumber\\
    &=\int\mathbf{s}_\theta(\mathbf{m}(\tau),\tau)^T\nabla_{\mathbf{m}(\tau)} \bigg[\int p(\mathbf{m}(\tau)|\mathbf{m}(0))p(\mathbf{m}(0))d\mathbf{m}(0)\bigg]d\mathbf{m}(\tau)\nonumber\\
    &=\int\mathbf{s}_\theta(\mathbf{m}(\tau),\tau)^T \bigg[\int p(\mathbf{m}(0))\nabla_{\mathbf{m}(\tau)} p(\mathbf{m}(\tau)|\mathbf{m}(0))d\mathbf{m}(0)\bigg]d\mathbf{m}(\tau).
\end{align}

Note that,
\begin{equation}
    \nabla_{\mathbf{m}(\tau)} p(\mathbf{m}(\tau)|\mathbf{m}(0))=\nabla_{\mathbf{m}(\tau)} \log p(\mathbf{m}(\tau)|\mathbf{m}(0))p(\mathbf{m}(\tau)|\mathbf{m}(0)),
\end{equation}
we develop Eq.~\ref{b4} as,
\begin{align}
    \int&\mathbf{s}_\theta(\mathbf{m}(\tau),\tau)^T\nabla_{\mathbf{m}(\tau)} p(\mathbf{m}(\tau))d\mathbf{m}(\tau)\nonumber\\
    &=\int\mathbf{s}_\theta(\mathbf{m}(\tau),\tau)^T \bigg[\int p(\mathbf{m}(0))p(\mathbf{m}(\tau)|\mathbf{m}(0))\nabla_{\mathbf{m}(\tau)} \log p(\mathbf{m}(\tau)|\mathbf{m}(0))d\mathbf{m}(0)\bigg]d\mathbf{m}(\tau)\nonumber\\
    &=\int \int p(\mathbf{m}(0))p(\mathbf{m}(\tau)|\mathbf{m}(0))\bigg[\mathbf{s}_\theta(\mathbf{m}(\tau),\tau)^T\nabla_{\mathbf{m}(\tau)} \log p(\mathbf{m}(\tau)|\mathbf{m}(0))\bigg]d\mathbf{m}(0)d\mathbf{m}(\tau)\nonumber\\
    &=\mathbb{E}_{p(\mathbf{m}(0))}\mathbb{E}_{p(\mathbf{m}(\tau)|\mathbf{m}(0))}[\mathbf{s}_\theta(\mathbf{m}(\tau),\tau)^T\nabla_{\mathbf{m}(\tau)} \log p(\mathbf{m}(\tau)|\mathbf{m}(0))].
\end{align}

Hence, the original score matching loss function is now expressed as,
\begin{align}\label{b7}
    \mathcal{L}^{\mathrm{SM}}(\theta)=
    &\mathbb{E}_{p(\tau)}\bigg[\lambda(\tau)\mathbb{E}_{p(\mathbf{m}(\tau))}\big[\|\mathbf{s}_\theta(\mathbf{m}(\tau),\tau)\|^2\big]\bigg] + C_1\mathbb{E}_{p(\tau)}[\lambda(\tau)] - \nonumber\\
    &\mathbb{E}_{p(\tau)}\bigg[\lambda(\tau)\mathbb{E}_{p(\mathbf{m}(0))}\mathbb{E}_{p(\mathbf{m}(\tau)|\mathbf{m}(0))}[\mathbf{s}_\theta(\mathbf{m}(\tau),\tau)^T\nabla_{\mathbf{m}(\tau)} \log p(\mathbf{m}(\tau)|\mathbf{m}(0))]\bigg].
\end{align}

The denoising score matching loss function is given by,
\begin{align}\label{b8}
    \mathcal{L}^{\mathrm{DSM}}(\theta)&=\mathbb{E}_{p(\tau)}\bigg[\lambda(\tau)\mathbb{E}_{p(\mathbf{m}(0))}\mathbb{E}_{p(\mathbf{m}(\tau)|\mathbf{m}(0))}\big[\|\mathbf{s_\theta}(\mathbf{m}(\tau),\tau)-\nabla_{\mathbf{m}(\tau)}\log p(\mathbf{m}(\tau) | \mathbf{m}(0)) \|^2\big]\bigg]\nonumber\\
    &=\mathbb{E}_{p(\tau)}\bigg[\lambda(\tau)\mathbb{E}_{p(\mathbf{m}(0))}\mathbb{E}_{p(\mathbf{m}(\tau)|\mathbf{m}(0))}\big[\|\mathbf{s_\theta}(\mathbf{m}(\tau),\tau)\|^2 + \|\nabla_{\mathbf{m}(\tau)}\log p(\mathbf{m}(\tau) | \mathbf{m}(0))\|^2\nonumber\\
    &\ \ \ \ \ -\mathbf{s_\theta}(\mathbf{m}(\tau),\tau)^T\nabla_{\mathbf{m}(\tau)}\log p(\mathbf{m}(\tau) | \mathbf{m}(0))\big]\bigg]\nonumber\\
    &=\mathbb{E}_{p(\tau)}\bigg[\lambda(\tau)\mathbb{E}_{p(\mathbf{m}(0))}\mathbb{E}_{p(\mathbf{m}(\tau)|\mathbf{m}(0))}\big[\|\mathbf{s}_\theta(\mathbf{m}(\tau),\tau)\|^2\big]\bigg] + C_2\mathbb{E}_{p(\tau)}[\lambda(\tau)] - \nonumber\\
    &\ \ \ \ \mathbb{E}_{p(\tau)}\bigg[\lambda(\tau)\mathbb{E}_{p(\mathbf{m}(0))}\mathbb{E}_{p(\mathbf{m}(\tau)|\mathbf{m}(0))}[\mathbf{s}_\theta(\mathbf{m}(\tau),\tau)^T\nabla_{\mathbf{m}(\tau)} \log p(\mathbf{m}(\tau)|\mathbf{m}(0))]\bigg]\nonumber\\
    &=\mathbb{E}_{p(\tau)}\bigg[\lambda(\tau)\mathbb{E}_{p(\mathbf{m}(\tau))}\big[\|\mathbf{s}_\theta(\mathbf{m}(\tau),\tau)\|^2\big]\bigg] + C_2\mathbb{E}_{p(\tau)}[\lambda(\tau)] - \nonumber\\
    &\ \ \ \ \mathbb{E}_{p(\tau)}\bigg[\lambda(\tau)\mathbb{E}_{p(\mathbf{m}(0))}\mathbb{E}_{p(\mathbf{m}(\tau)|\mathbf{m}(0))}[\mathbf{s}_\theta(\mathbf{m}(\tau),\tau)^T\nabla_{\mathbf{m}(\tau)} \log p(\mathbf{m}(\tau)|\mathbf{m}(0))]\bigg],
\end{align}
based on the definition that $\|\nabla_{\mathbf{m}(\tau)}\log p(\mathbf{m}(\tau) | \mathbf{m}(0)) \|^2\coloneqq C_2$ is a constant independent of $\theta$.

Looking at Eqs.~\ref{b7} and~\ref{b8}, we see that,
\begin{equation}
    \mathcal{L}^{\mathrm{DSM}}(\theta)=\mathcal{L}^{\mathrm{SM}}(\theta)+\tilde{C},
\end{equation}
where
\begin{equation}
    \tilde{C}=C_2\mathbb{E}_{p(\tau)}[\lambda(\tau)]-C_1\mathbb{E}_{p(\tau)}[\lambda(\tau)],
\end{equation}
is also a constant independent of $\theta$ and does not affect the optimization process. Therefore, we conclude that $\mathcal{L}^{\mathrm{DSM}}$ is equivalent to $\mathcal{L}^{\mathrm{SM}}$ up to a constant.




\section*{Acknowledgments}
This work is supported by the National Key Research and Development Project, China (No. 2023YFE0110900), National Natural Science Foundation of China (Nos. 42320104003, 42077247), Tencent CarbonX Program, and CCF-Tencent Rhino-Bird Open Research Fund. Z.F. would like to thank the International Exchange Program for Graduate Students at Tongji University.

\section*{Data availability}
All the data and codes needed to evaluate the conclusions in the paper will be made publicly available on \url{https://github.com/fengzhao1239/DiffNO} upon acception.

\section*{Compliance with ethical standards}
Conflict of Interest: The authors declare that they have no conflict of interest.

\bibliographystyle{elsarticle-num-names}
\bibliography{refs}

\end{document}